\documentclass[lettersize,journal]{IEEEtran}
\usepackage{amsmath,amsfonts}
\usepackage{algorithmic}
\usepackage{algorithm}
\usepackage{array}
\usepackage[caption=false,font=normalsize,labelfont=sf,textfont=sf]{subfig}
\usepackage{textcomp}
\usepackage{stfloats}
\usepackage{url}
\usepackage{verbatim}
\usepackage{graphicx}
\usepackage{cite}
\usepackage{tikz}
\usetikzlibrary{arrows.meta,positioning,calc}
\usepackage{booktabs}
\usepackage{tabularx}
\usepackage{longtable}
\usepackage{enumitem}
\usepackage{multirow}
\setlist{topsep=2pt,itemsep=1pt,parsep=0pt}
\AtBeginDocument{%
  \setlength{\textfloatsep}{6pt plus 2pt minus 2pt}%
  \setlength{\floatsep}{6pt plus 2pt minus 2pt}%
  \setlength{\intextsep}{6pt plus 2pt minus 2pt}%
  \setlength{\abovecaptionskip}{3pt}%
  \setlength{\belowcaptionskip}{2pt}%
}
\begin{document}

\title{Knowledge-Conditioned, Single-Pass LLM Synthesis of Executable Unity Game Scenes: A Compiler Error Census across 26 Goal Playable Concepts}

\author{Hugh~Xuechen~Liu and K{\i}van\c{c}~Tatar%
\thanks{Hugh Xuechen Liu and K{\i}van\c{c} Tatar are with Chalmers University of Technology and University of Gothenburg, SE-412~96 G\"oteborg, Sweden (e-mail: xuechen@chalmers.se; tatar@chalmers.se).}}


\maketitle

\begin{abstract}
Large language models (LLMs) write Unity C\# for game scenes. Yet nearly all demonstrations rest on an iterative repair loop that regenerates code until it compiles, conflating what the model writes with what the loop fixes. We remove the loop and evaluate a single pass, where the first draft is final. This isolates the model's parametric knowledge, the most stringent test of unaided generation. Models instantiate Goal Playable Concepts, playable counterparts of goal patterns, across 10,400 generations (four open-weight models, 7B--30B; two generation modes; four intermediate-representation (IR) conditioning levels; 26 goal patterns; 20 seeds). None compiled into a runnable scene, leaving no survivorship bias. To understand how the generated C\# scripts fail, we categorize the 99 error codes behind 90{,}673 compiler-error occurrences as Grounding (invented or misused Unity types and APIs) or Hygiene (structural defects needing no Unity knowledge). The split differs sharply by goal pattern (e.g., Stealth fails mostly on invented engine references; Capture on plain C\# structure). Larger models, stricter IRs, and different generation modes move the errors but never yield a compiling scene. The bottleneck is missing engine-specific knowledge. The census orders goal patterns by that demand, showing designers where single-pass generation breaks.
\end{abstract}


\section{Introduction}
\label{sec:intro}

\IEEEPARstart{L}{arge} language models (LLMs) have become a practical tool for game content generation. Demos and tutorials routinely show LLMs writing Unity C\# scripts, placing game objects, and implementing mechanics on demand~\footnote{Some examples: \url{https://www.youtube.com/watch?v=gSFHyso_uuI}; \url{https://github.com/keijiro/DungeonMatchHeroes}}. Yet almost every such demonstration rests on an \emph{iterative repair loop}: the model generates a candidate, the compiler errors (if any) return to a human or to the model itself, and the loop repeats until the artifact compiles and runs. This workflow is productive in practice but conflates what the model writes for the scene with what the repair loop fixes. Also, the gameplay on show is likewise picked ad hoc rather than drawn from a gameplay design vocabulary~\cite{church1999formal,costikyan2002have,kreimeier2002case,cairns2019future}.

We ask a simpler and harder question: \emph{what is the intrinsic capability ceiling of single-pass LLM generation of executable game artifacts with no human feedback and no iterative repair?} Single-pass evaluation isolates the model's parametric knowledge (what is stored in its weights) from the practitioner's domain expertise. It is the most stringent and most diagnostic condition. In an iterative workflow, each error reflects the model plus the feedback it received. Removing the loop makes every error attributable to the model alone. Removing the loop also levels the comparison. Every model faces the same condition on every task. A difference in the error profile can be hence read as a difference in what the model knows or in what the task demands.

Existing game generation studies typically condition on genre labels (e.g., First-person Shooter, platformer) or free-form natural language. However, genre is criticised as a culturally constructed, semantically unstable concept that cannot serve as a reproducible evaluation target for cumulative research~\cite{clarke2017video,arsenault2009video,juul2011half}. We instead ground our evaluation in \emph{Goal Playable Concepts} (GPC)~\cite{lyu_gpc_2023}, distinguished across three layers. The gameplay design pattern language of Bj\"ork and Holopainen~\cite{bjork2005patterns} is a theoretical design vocabulary following the ontological and formal analysis approach of game research~\cite{lankoski2015game}. A pattern means a frequently recurring configuration in gameplay design. Within it, 26 \emph{goal patterns} describe what a player is trying to achieve. A GPC is the playable counterpart of one goal pattern, an artifact built so the concept can be grasped by playing it. Each GPC in turn has a reference \emph{Unity instantiation} (currently its only engine realization so the two are coextensive here). This layering makes GPC the right evaluation target, giving a theoretically grounded abstraction, an executable artifact, and an objective binary verdict from compilation across all 26 goal patterns.

Our evaluation spans five [model, generation mode], where the generation mode is Editor-style or Runtime-builder (defined in Section~\ref{sec:design}). Four open-weight models, each from a different family, generate in the Editor-style mode (7B-Qwen2.5, 16B-DeepSeek, 22B-Codestral, 30B-Qwen3-Ed). The largest additionally generates in the Runtime-builder mode (30B-Qwen3-Rt). Each [model, generation mode] runs the identical grid of 4 intermediate-representation (IR) conditioning levels, 26 goal patterns, and 20 random seeds. This yields \textbf{10,400 generation records}, each a single generation pass producing one Unity C\# script with no repair. Open-weight models are a deliberate choice. They deploy locally (which studios with confidential assets require), are reproducible (frozen checkpoints do not drift like commercial endpoints), and admit fine-tuning at realistic cost. The 7B--30B range covers the tier these constraints admit.

We find that \emph{no generated C\# script compiled into a runnable scene across any of the 10,400 records}. We count a record as successful only when it both compiles and exposes a valid Unity entry point (Section~\ref{sec:design} details the outcome taxonomy). Rather than a verdict, this uniform outcome is what makes the dataset suited to exhaustive failure-mode enumeration. Every record contributes failure evidence, and no survivorship filter separates analysed from discarded attempts.

\textbf{Contributions.} This paper makes four contributions. \textbf{(1) A Grounding/Hygiene taxonomy for LLM-generated Unity C\# scripts.} Drawing on the domain-specific versus domain-independent error distinction from software-engineering and computing-education research, we categorize all 99 observed C\# error codes as \emph{Grounding} (missing or incorrect Unity API/type knowledge) or \emph{Hygiene} (domain-independent structural defects), comprising 18 Grounding and 81 Hygiene error codes. \textbf{(2) A 10,400-record error code census with per-pattern profiles.} We count every error-code occurrence across the 10,400 records (90,673 total across 99 error codes). The per-pattern Grounding share ranges from 0 to 0.98 (a property of error composition, robust to the error-volume inflation of 22B-Codestral reported in Section~\ref{sec:results}), indicating that goal patterns surface qualitatively different regions of the model's Unity API knowledge which we interpret rather than directly measure as a knowledge boundary, and visualise as a Grounding--Hygiene scatter across the four IR levels. \textbf{(3) An error-derived lens on goal-pattern semantic complexity.} Linking these profiles to each pattern's game design semantics yields an empirical ordering of Bj\"ork and Holopainen's goal patterns~\cite{bjork2005patterns} by the engine knowledge their instantiation demands. Perception- and physics-coupled patterns (e.g., \emph{Stealth}, \emph{Rescue}) concentrate Grounding errors, whereas patterns reducible to general state manipulation (e.g., \emph{Capture}) concentrate Hygiene errors. This connects a design-pattern vocabulary to an observable error structure of playable-scene synthesis and shows designers which gameplay concepts current LLMs struggle to realize. \textbf{(4) Condition-axis observations: IR conditioning and generation mode.} Across the four IR levels, per-pattern profiles shift in both magnitude and Grounding/Hygiene composition, and a 30B Editor-versus-Runtime comparison shows that switching generation mode (editor versus runtime, Section~\ref{sec:design}) redistributes rather than resolves Grounding errors, locating the bottleneck in the model's missing engine knowledge rather than generation mode \footnote{This study is reproducible through \url{https://anonymous.4open.science/r/paper-repo-failure-taxonomy-LLM-Unity-Scene-2D47/README.md}}.

The remainder of this paper is organised as follows. Section~\ref{sec:background} situates the work relative to prior LLM code generation, game content generation, and compiler error taxonomy literature. Section~\ref{sec:design} describes the experimental design, IR conditioning protocol, and failure taxonomy construction, and walks through one pattern end to end as a concrete exemplar. Section~\ref{sec:results} reports results. Section~\ref{sec:discussion} interprets the findings and discusses limitations. Section~\ref{sec:conclusion} concludes.

\section{Background}
\label{sec:background}

This section situates our study against three literatures. We first review how LLM code generation is evaluated since our single-pass, compile-or-fail setup departs from the typical function-level benchmarks. We then turn to LLM-assisted game content generation to position goal playable concepts against the genre-based and natural-language-based targets used in prior work. We finally draw on the compiler-error taxonomy tradition in computing education which supplies the domain-dependent versus domain-independent distinction that our ``Grounding/Hygiene'' split builds on.

\subsection{LLM Code Generation Evaluation}

Following prior work that contrasts function-, project-, and library-level code generation~\cite{chen2021evaluating,austin2021program,rahman2025beyond,jiang2026survey}, we distinguish three settings: \emph{function-level} tasks on self-contained snippets, \emph{project-level} generation that must fit into a multi-file codebase, and \emph{library-level} generation that must call a specific external API correctly. Difficulty rises as code leaves the self-contained setting. Function-level benchmarks with automated test suites are the current dominant evaluation setting for LLM code generation. HumanEval~\cite{chen2021evaluating} and MBPP~\cite{austin2021program} each provide hundreds of self-contained Python programming problems with accompanying tests. Performance is reported as $\text{pass@}k$, the probability that at least one of $k$ independently sampled completions passes all hidden unit tests. These benchmarks have driven rapid progress but assess a narrow profile, namely isolated functions with no external type dependencies.

Difficulty rises as code leaves the self-contained setting in function-level tasks (as demonstrated before) which need no external types. For example, project-level generation must fit a multi-file codebase. Also, library-level generation must call a specific external API correctly with the right type names, signatures, and namespaces. Once the artifact depends on such an API, the model's parametric knowledge of that API might become the limiting factor. Studies on domain-specific libraries and real-world repositories find accuracy degrades sharply relative to general benchmarks, as models conflate API versions, invent non-existent method names, or reference types absent from the target library~\cite{jiang2026survey,jimenez2023swe}. Engine-coupled generation, our setting, must bind to a live (game) engine scene rather than a static library. It remains largely unexamined.

\subsection{LLM-Assisted Game Content Generation}

Procedural content generation (PCG) has a long history in games, spanning level layout, terrain, narrative quest design, and item parametrisation~\cite{shaker2016procedural}. PCG has established search-based~\cite{togelius2011sbpcg} and machine-learning~\cite{summerville2018pcgml} traditions. LLMs have more recently been applied to tasks once handled by explicit generators such as rule-based narrative systems, grammar-driven level generators, and constraint solvers~\cite{gallotta2024llmgames,todd2023level,sudhakaran2023mariogpt}. Much of this work focuses on genre labels (e.g., platformer, Role-Play Games) or free-form descriptions. However, genre is a coarse, surface-level label applied to a whole game. By contrast, game content and design form a deeper and structured space. Hendrikx et al.~\cite{hendrikx2013procedural} organise that space into six layers ordered by complexity, running from low-level game bits through game space, game systems, and game scenarios up to the game-design layer of rules and \emph{goals} that the lower layers serve. Vocabularies like game design pattern~\cite{bjork2005patterns,barney2020pattern} and game ontologies~\cite{zagal2007towards,parkkila2017ontology,de2023modeling} sit at this game-design layer rather than at the genre surface. That depth makes them a more principled evaluation target than a genre tag. Prior game research further criticises genre as culturally constructed and semantically unstable, too unstable to serve as a reproducible evaluation target for cumulative research~\cite{clarke2017video,arsenault2009video,juul2011half}.

Among the forementioned game design patterns and ontologies, Goal Playable Concepts (GPC)~\cite{lyu_gpc_2023} provide a theoretically grounded alternative. Within Bj\"ork and Holopainen's gameplay design pattern language~\cite{bjork2005patterns}, \emph{goal patterns} describe player goals and their interactive affordances (listed in Table~\ref{tab:patterns}), and a GPC realizes one goal pattern as a playable artifact (currently a verifiable Unity scene). This layering enables evaluation at the gameplay pattern level rather than the genre level, and makes failure analysis directly interpretable in terms of design-concept structure. Structured, executable game representations also underpin frameworks such as the Video Game Description Language ~\cite{ebner2013towards} and GVGAI~\cite{perezliebana2019gvgai} which target agent play rather than generation from design concepts.

To our knowledge, no prior game-content work has compared failure profiles systematically at the level of individual game design patterns. Closest to our setting, the contemporaneous Mage benchmark~\cite{mage2026anon} evaluates LLM-generated executable game scenes along multiple quality axes on a related class of goal-pattern tasks. The two efforts are complementary and differ in their unit of analysis. Mage measures \emph{how well} generation succeeds under benchmark metrics, whereas the present study measures \emph{how} generation fails at the granularity of individual compiler error occurrences. Understanding how these errors distribute across pattern instantiations and model configurations matters for two reasons. First, it locates where an LLM's scene-generation capability breaks down. Second, it shows how the semantics of each goal pattern constrains what its Unity instantiation must implement.

\begin{table}[!t]
\renewcommand{\arraystretch}{1.15}
\caption{The 26 goal patterns and their one-line descriptions.}
\label{tab:patterns}
\centering
\footnotesize
\begin{tabular}{@{}l p{5.6cm}@{}}
\hline
\bfseries Pattern & \bfseries One-line Description \\
\hline
Ownership          & Gain ownership of a game element. \\
Collection         & Complete several goals that together form a coherent unit. \\
Eliminate          & Remove a game element from its location in the game space. \\
Capture            & Eliminate or take ownership of an actively resisting goal object. \\
Overcome           & Defeat an opposing force in a test or series of tests. \\
Evade              & Avoid being captured or hit. \\
Stealth            & Move through an area and act without being detected. \\
Herd               & Move a game element to a location without directly interacting with it. \\
Conceal            & Hinder other players' ability to gain information. \\
Rescue             & Free someone or something that is guarded. \\
Delivery           & Move a game element to a specified element or place in the game space. \\
Guard              & Hinder others from accessing a particular area or game element. \\
Race               & Be the first to reach a goal, often a location via an approved route. \\
Alignment          & Form a linear alignment of game elements. \\
Configuration      & Form a spatial, temporal, or logical arrangement of game elements. \\
Traverse           & Move a game element from one position in the game to another. \\
Survive            & Avoid being killed by other players' actions and game events. \\
Connection         & Link or position game elements so they have a physical relation. \\
Exploration        & Learn the layout of the game world, or locate parts or objects in it. \\
Reconnaissance     & Patrol a known area to detect changes. \\
Contact            & Bring two or more elements into physical contact. \\
Enclosure          & Surround game elements with a continuous line or wall. \\
Gain Competence    & Gain the ability to perform a certain action in the game. \\
Gain Information   & Act in the game to receive information or make deductions. \\
Last Man Standing  & Be the last survivor. \\
King of the Hill   & Reach and keep a sought-after game state that others also want. \\
\hline
\end{tabular}
\end{table}

\subsection{Compiler Error Taxonomies}

Compilation is a strict and objective gate. A scene either builds and runs or it does not with no rubric or human judgement in between. This makes compile success or failure an unusually clean signal for an executable artifact. Unlike open-ended text, a failed compile is unambiguous and its cause is localised to a specific diagnostic. The type of compiler error is itself diagnostic of why the model failed. A missing semicolon and an invented engine type are both failures, but they implicate very different model capabilities. We treat the compiler not as a mere pass-or-fail oracle but as a structured probe into the model's failure modes. The following taxonomy formalises that probe. Throughout the paper, \emph{failure} refers to the outcome of a generation attempt that yields no runnable scene. \emph{Error} refers to a compiler diagnostic that such an attempt emits.

Classifying compiler errors by semantic type is well established in computer science education research. Altadmri and Brown's analysis of the Blackbox dataset identified characteristic error distributions for novice programmers, showing that a small set of error types accounts for the majority of all compile-time failures and that certain errors (e.g., missing semicolons, unmatched braces) are structurally domain-independent while others reflect gaps in conceptual understanding~\cite{altadmri2015compilations}. Pettit et al.\ and subsequent studies confirm this concentration effect and connect error type to the underlying cognitive difficulty of the corresponding programming concept~\cite{pettit2017enhanced}.

This domain-independent versus domain-dependent distinction is the conceptual basis we build on. For engine-coupled generation the domain-dependent side becomes engine-specific. Some errors cannot be resolved without knowing what types and methods the target engine actually provides, whereas others (e.g., a missing semicolon, an unmatched brace) are domain-independent and can be diagnosed with no engine knowledge. We do not assume such a taxonomy a priori. Section~\ref{sec:results} derives one from the error codes actually observed across the 10,400 records.

Prior work has used error type profiles to characterize LLM code quality beyond $\text{pass@}k$, identifying that specific error categories concentrate in predictable model failure modes~\cite{tambon2024bugs,dou2024whats}. We extend this approach to game content generation, connecting per-code frequency profiles to the semantic complexity of the underlying goal pattern, a mapping that has not been made in prior game generation or LLM evaluation research.

\section{Research Design}
\label{sec:design}

\subsection{Task Definition}


We evaluate whether open-weight LLMs can instantiate the 26 GPC~\cite{lyu_gpc_2023} (introduced in Section~\ref{sec:intro}) as executable Unity C\# scripts in a \emph{single generation pass}, with no human feedback and no iterative repair. We use ``single-pass'' in a specific sense: the absence of an error-driven repair loop that feeds compiler errors back into the code. The \texttt{with\_schema} runs call the model twice, once for the IR and once for the code. Neither call sees compiler output. We never re-run a generation after it fails. The two calls are two steps of one pipeline, not two rounds of fixing. Single-pass is not one-shot or few-shot, terms that count the worked examples placed in the prompt~\cite{Ramesh2025ZeroFewShotNLP}. Our prompts contain none and are zero-shot in that sense. This single-pass condition isolates the model's own knowledge from a practitioner's debugging skill, exposing failure modes that iterative workflows conceal.

Every GPC has a reference Unity instantiation. These references are our proof of feasibility only. Generation runs inside a single pre-built Unity project holding all 26 scenes and their shared assets. The model sees none of this: not the reference instantiations, not the project contents. The task is to produce a complete and self-contained artifact. The script must create whatever it needs (e.g., game objects, components, and materials) through Unity's programmatic APIs. The generated script is compiled inside the project. A reference to an asset that already exists under its correct name resolves against it. Anything else the script must create itself. A reference to a type or asset that neither exists nor is created becomes a compiler error.

\begin{figure}[!t]
\centering
\resizebox{\linewidth}{!}{%
\begin{tikzpicture}[
  node distance=2mm and 3mm,
  box/.style={draw, rounded corners=1pt, align=center, minimum height=5mm, minimum width=11mm, inner sep=1pt, font=\footnotesize},
  model/.style={box, fill=gray!10},
  data/.style={box},
  schema/.style={box, dashed, fill=gray!5},
  artifact/.style={box, fill=gray!20},
  >={Stealth[length=1.4mm]},
  flow/.style={->, >=Stealth, shorten >=1pt, shorten <=1pt, thick},
  inject/.style={->, >=Stealth, dashed, shorten >=1pt, shorten <=1pt},
]
\node[data] (B1) {pattern\_md};
\node[model, right=of B1] (B2) {LLM};
\node[data, right=of B2] (B3) {C\#};
\node[artifact, right=of B3] (B4) {Unity};
\node[data, right=of B4] (B5) {err log};
\node[left=1mm of B1, font=\footnotesize\itshape, anchor=east] {Base};
\draw[flow] (B1) -- (B2); \draw[flow] (B2) -- (B3); \draw[flow] (B3) -- (B4); \draw[flow] (B4) -- (B5);
\node[data, below=6mm of B1] (I1) {pattern\_md};
\node[model, right=of I1] (I2) {LLM$_1$};
\node[data, right=of I2] (I3) {IR};
\node[model, right=of I3] (I4) {LLM$_2$};
\node[data, right=of I4] (I5) {C\#};
\node[artifact, right=of I5] (I6) {Unity};
\node[data, right=of I6] (I7) {err log};
\node[left=1mm of I1, font=\footnotesize\itshape, anchor=east] {IR-cond};
\draw[flow] (I1) -- (I2); \draw[flow] (I2) -- (I3); \draw[flow] (I3) -- (I4); \draw[flow] (I4) -- (I5); \draw[flow] (I5) -- (I6); \draw[flow] (I6) -- (I7);
\node[schema, below=5mm of I2] (S) {schema\\(\texttt{free}/\texttt{min}/\texttt{full})};
\draw[inject] (S) -- (I2);
\end{tikzpicture}%
}
\caption{Baseline and IR-conditioned pipelines. Data (white), LLM (light grey), compile target (dark grey), schema injection (dashed). Both share the same Unity batch-replay harness and error-log aggregation.}
\label{fig:pipeline}
\end{figure}
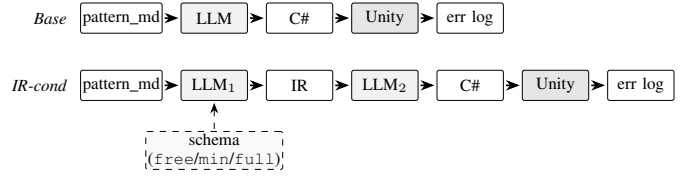

We use two generation pipelines (Figure~\ref{fig:pipeline}). The \texttt{no\_schema} pipeline gives the model only the goal pattern description, a markdown document defining the pattern's game-design semantics, and asks for a Unity C\# script directly. The \texttt{with\_schema} pipeline adds an intermediate step. Step~1 generates an intermediate representation (IR) in JSON from the pattern description and a schema document. Step~2 has the same model generate the C\# script from its own IR. The reference IRs that accompany the reference scenes are never used. The model builds its own IR, then implements it. The IR has seven top-level fields (\texttt{scene}, \texttt{objects}, \texttt{scripts}, \texttt{params}, \texttt{runtime\_params}, \texttt{links}, \texttt{rules}). The \texttt{with\_schema} pipeline runs at three levels (\texttt{free}, \texttt{min}, \texttt{full}). They differ in how much of the frozen schema the model sees. More schema means more field-level constraints to satisfy. \texttt{free} gives no schema text, leaving the model to structure the IR itself. \texttt{min} gives only the top-level field names. \texttt{full} gives the complete field definitions and hard constraints (Appendix~\ref{app:ir}).

This IR schema is fixed before the whole experiment. Building it was an offline design step by the authors. We started from three hand-authored reference implementations. Over five versioned iterations we revised the IR schema into \emph{v0.2-runtime-evidence}, then froze it before generation began. Freezing keeps the IR axis uniform across runs, not tuned to outcomes. This offline design does not weaken the single-pass condition. The model still writes every script in one pass with no compiler feedback. A field-frequency check across all 26 reference implementations confirms that every GPC populates all seven fields. The details of the IR schema's design and iteration are also provided in Appendix~\ref{app:ir}.

\subsection{Experimental Variables}

Table~\ref{tab:factors} lists all experimental factors. Because the target engine is Unity throughout, we first fix a distinction that Unity itself draws. A scene can be built along two different code paths. Our two Unity script generation modes correspond to them. The \emph{Editor-style} mode requests a Unity Editor script that builds the scene at edit time through editor-side APIs (e.g., the \texttt{UnityEditor} namespace, \texttt{AssetDatabase}, menu actions).\footnote{Unity Editor scripting API: \url{https://docs.unity3d.com/ScriptReference/AssetDatabase.html}, \url{https://docs.unity3d.com/ScriptReference/MenuItem.html}.} The \emph{Runtime-builder} mode requests a single MonoBehaviour that constructs the scene programmatically in \texttt{Awake()} at play time, defines all gameplay classes in the same file, and may not use the \texttt{UnityEditor} namespace at all.\footnote{Unity runtime scripting API: \url{https://docs.unity3d.com/ScriptReference/MonoBehaviour.html}, \url{https://docs.unity3d.com/ScriptReference/MonoBehaviour.Awake.html}.} The two paths exercise different parts of the Unity API. Contrasting them at a fixed model tests whether failures are tied to one API regime or intrinsic to the model's engine knowledge.

The five [model, generation mode] come from crossing four pre-trained and frozen-weight open models with the two generation modes. Three models generate in the Editor-style mode only: Qwen2.5-Coder-7B~\cite{hui2024qwen2}, DeepSeek-Coder-V2-Lite-16B~\cite{zhu2024deepseek}, and Codestral-22B~\footnote{\url{https://huggingface.co/mistralai/Codestral-22B-v0.1}} (7B-Qwen2.5, 16B-DeepSeek, 22B-Codestral). The fourth, Qwen3-Coder-30B~\cite{qwen3technicalreport} (a newer generation than the Qwen2.5-Coder used at 7B), generates in both modes (30B-Qwen3-Ed and 30B-Qwen3-Rt). Switching models changes identity as a whole, not scale alone. Size, family, and pretraining corpus change together. The 30B Editor-versus-Runtime contrast holds identity fixed and isolates the generation mode. A still larger or newer open-weight model (e.g., a Gemma-class checkpoint) is the same scale remedy at greater magnitude. Checkpoints released after this study are a replication target, not an omission.

Four IR conditioning levels vary the structured context provided. At one end \texttt{no\_schema} supplies only the pattern description. At the other \texttt{with\_schema\_full} adds a complete GPC schema with all field constraints. Twenty seeds per [[model, generation mode], condition, pattern] capture run-to-run variation.

The factor grid of 26 goal patterns $\times$ 4 IR conditioning levels $\times$ 20 seeds is identical for every [model, generation mode], giving 2,080 records each and 10,400 in total (Table~\ref{tab:factors}).

\begin{table}[!t]
\renewcommand{\arraystretch}{1.3}
\caption{Experimental factors. Each of the five [model, generation mode] runs the identical IR-by-pattern-by-seed grid for 2{,}080 records (10{,}400 total).}
\label{tab:factors}
\centering
\begin{tabular}{p{1.4cm}p{4.0cm}p{2.0cm}}
\hline
\bfseries Factor & \bfseries Levels & \bfseries Notes \\
\hline
[Model, generation mode] &
  7B-Qwen2.5, 16B-DeepSeek, 22B-Codestral, 30B-Qwen3-Ed, 30B-Qwen3-Rt &
  5 [model, generation mode] \\[2pt]
IR conditioning &
  \texttt{no\_schema}, \texttt{with\_schema\_free},
  \texttt{with\_schema\_min}, \texttt{with\_schema\_full} &
  Increasing schema detail \\[2pt]
Goal pattern & 26 goal patterns & Unity inst.\ for all \\[2pt]
Seed & 20 per [[model, generation mode], condition, pattern] & Stochastic variation \\
\hline
\end{tabular}
\end{table}

\begin{figure}[!t]
\centering
\resizebox{\linewidth}{!}{%
\begin{tikzpicture}[
  node distance=2mm and 2.5mm,
  cell/.style={draw, rounded corners=1pt, align=center, inner sep=2pt, font=\footnotesize, minimum height=11mm},
  phase/.style={cell, fill=gray!10},
  shared/.style={cell, dashed, fill=gray!5},
  grouplab/.style={font=\footnotesize\itshape},
  >={Stealth[length=1.4mm]},
  flow/.style={->, >=Stealth, shorten >=1pt, shorten <=1pt},
]
\node[phase] (c1) {\textbf{7B-Qwen2.5}\\Qwen2.5-Coder\\Editor mode};
\node[phase, right=of c1] (c2) {\textbf{16B-DeepSeek}\\DeepSeek-V2-Lite\\Editor mode};
\node[phase, right=of c2] (c3) {\textbf{22B-Codestral}\\Codestral\\Editor mode};
\node[phase, right=of c3] (c4) {\textbf{30B-Qwen3-Ed}\\Qwen3-Coder\\Editor mode};
\node[phase, right=of c4] (c5) {\textbf{30B-Qwen3-Rt}\\Qwen3-Coder\\Runtime mode};
\node[grouplab, above=1.5mm of c2] {Model axis (fixed mode)};
\node[grouplab, above=1.5mm of $(c4.north)!0.5!(c5.north)$] {Mode axis (fixed model)};
\node[shared, below=4mm of c3, minimum height=6mm] (grid)
  {shared factor grid per [model, generation mode]: 26 goal patterns $\times$ 4 IR levels $\times$ 20 seeds $=$ 2{,}080 records};
\foreach \c in {c1,c2,c3,c4,c5} \draw[flow] (grid) -- (\c);
\end{tikzpicture}%
}
\caption{The five [model, generation mode] and the shared factor grid. Four open-weight models generate in the Editor-style mode. The largest also generates in the Runtime-builder mode. Each runs the identical factor grid (bottom). Total: $5 \times 2{,}080 = 10{,}400$ records.}
\label{fig:phases}
\end{figure}
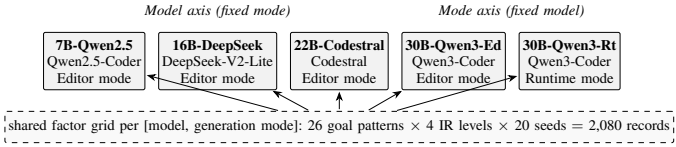

Throughout the paper, we write the design axes as tuples. The five [model, generation mode] are fixed pairings, not a full model-by-mode crossing. We bracket the pair as one unit inside larger tuples such as [[model, generation mode], condition, pattern, seed]. A [model, generation mode] is one model generating in one mode, Editor-style or Runtime-builder (five: 7B-Qwen2.5, 16B-DeepSeek, 22B-Codestral, 30B-Qwen3-Ed, 30B-Qwen3-Rt). A \emph{condition} is shorthand for an IR conditioning level (four: \texttt{no\_schema}, \texttt{free}, \texttt{min}, \texttt{full}). A \emph{record} is one generation attempt for a given [[model, generation mode], condition, pattern, seed]. Figure~\ref{fig:phases} summarises the five [model, generation mode] and the shared factor grid.

\subsection{Generation and Compilation Pipeline}
\label{subsec:pipeline}

Each record comes from one [[model, generation mode], condition, pattern, seed] (Table~\ref{tab:factors}). The model generates a Unity C\# script. The script is compiled. … The result is stored (i.e., the generated C\# script, its compilation outcome, and any emitted diagnostics). All [model, generation mode] generate with temperature 0.2 and top-p 0.95 on vLLM, under an output budget of 2,048 tokens for 7B-Qwen2.5 and 16B-DeepSeek and 8,192 tokens for 22B-Codestral and both 30B modes. These settings follow common practice rather than task-specific tuning. The output budgets are rarely reached and do not affect the results or their interpretation.\footnote{Each budget is the cap on generated tokens, within the model's context window (3,072 tokens for 7B-Qwen2.5 and 16B-DeepSeek, 32,768 for the others). Measured with each model's tokenizer, outputs reach the budget in 4.0\% of smaller-model records overall, 1.8\% under schema conditioning, and 2.1\% in 30B-Qwen3-Ed. The exceptions are the \texttt{no\_schema} runs of 7B-Qwen2.5 and 30B-Qwen3-Rt (16.9\% and 17.3\%), where free-form generation runs long and error counts include a truncation contribution. 22B-Codestral is a separate case, with incomplete outputs that stop far below its budget for reasons given in Section~\ref{sec:discussion}.} Each record follows an automated pipeline:

\begin{enumerate}
\item The LLM generates a Unity C\# script (and in the \texttt{with\_schema} conditions, first generates an IR JSON in a preceding call).
\item A format \emph{sanitizer} checks the raw output before it reaches the project. Records that fail this check (e.g., output that is empty after code-fence extraction, or that contains no C\# type declaration) are recorded as \emph{sanitizer rejected} and do not proceed to compilation.
\item A BatchRunner process writes the surviving script to the Unity project and triggers \texttt{AssetDatabase.Refresh} which imports and compiles the new \texttt{.cs} file and reloads the scripting domain.
\item Script compilation runs asynchronously relative to asset import. A watchdog enforces a 120-second timeout on the compile-and-domain-reload cycle, uniformly across every [model, generation mode] and IR conditioning level. Output is captured and parsed for C\# diagnostic codes. A record with captured diagnostics is a \emph{compile error}. A record where the timeout elapses before any diagnostics appear is \emph{timeout, no diagnostics}.
\item If compilation succeeds, BatchRunner checks for a recognised entry point. A script with a valid entry point is a \emph{pass}. One that compiles but exposes no valid entry point is \emph{no entry} (discussed in Section~\ref{sec:discussion}).
\item Two artifact codes are excluded from the census: \texttt{CS2001} is a start-up artifact emitted before the \texttt{AICommandRunStart} log marker and is never counted; \texttt{CS1029} lines containing the string \texttt{BatchRunner\_sanitize} mark a sanitizer rejection and are tallied separately from compilation failures.
\end{enumerate}

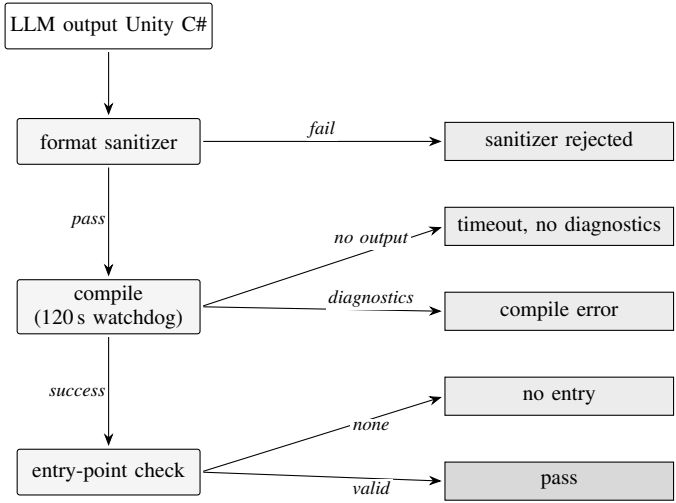
\begin{figure}[!t]
\centering
\resizebox{\linewidth}{!}{%
\begin{tikzpicture}[
  snode/.style={draw, rounded corners=1pt, align=center, font=\footnotesize,
               minimum height=6mm, minimum width=24mm, inner sep=2pt},
  dnode/.style={snode, fill=gray!8},
  leaf/.style={draw, align=center, font=\footnotesize,
              minimum height=5mm, minimum width=30mm, inner sep=2pt},
  lfail/.style={leaf, fill=gray!15},
  lpass/.style={leaf, fill=gray!30},
  >={Stealth[length=1.8mm]},
  flow/.style={->, >=Stealth, shorten >=1pt, shorten <=1pt},
  lbl/.style={font=\scriptsize\itshape, fill=white, inner sep=1pt},
]
\node[snode] (gen) {LLM output Unity C\#};
\node[dnode, below=9mm of gen] (san) {format sanitizer};
\node[dnode, below=15mm of san] (comp) {compile\\(120\,s watchdog)};
\node[dnode, below=15mm of comp] (entry) {entry-point check};
\node[lfail, right=32mm of san] (rej) {sanitizer rejected};
\node[lfail, below=6mm of rej] (to) {timeout, no diagnostics};
\node[lfail, below=6mm of to] (ce) {compile error};
\node[lfail, below=6mm of ce] (ne) {no entry};
\node[lpass, below=6mm of ne] (ps) {pass};
\draw[flow] (gen) -- (san);
\draw[flow] (san) -- node[lbl,left]{pass} (comp);
\draw[flow] (comp) -- node[lbl,left]{success} (entry);
\draw[flow] (san.east) -- node[lbl,above]{fail} (rej.west);
\draw[flow] (comp.east) -- node[lbl,above,pos=0.7]{no output} (to.west);
\draw[flow] (comp.east) -- node[lbl,above,pos=0.7]{diagnostics} (ce.west);
\draw[flow] (entry.east) -- node[lbl,below,pos=0.7]{none} (ne.west);
\draw[flow] (entry.east) -- node[lbl,below,pos=0.7]{valid} (ps.west);
\end{tikzpicture}%
}
\caption{Outcome routing for a single record. Each record enters at the top and exits at exactly one of the five mutually exclusive outcomes of Table~\ref{tab:outcomes} (grey, right). \emph{Pass} (compiled with a valid entry point) was never reached in any of the 10,400 records.}
\label{fig:outcomes}
\end{figure}

Figure~\ref{fig:outcomes} summarizes the generation and compilation pipeline. Each record falls into exactly one of the five mutually exclusive outcomes of Table~\ref{tab:outcomes}. A timeout does not discard a record. All diagnostics emitted before the 120-second limit are captured. Timed-out records with captured output are a primary data source for the error census (Section~\ref{sec:results}).\footnote{Because the 120-second compile limit is identical across every [model, generation mode] and IR conditioning level, cross-condition comparisons are not confounded by unequal timeouts. The limit is a heuristic, long enough for the compile-and-domain-reload cycle to finish in typical cases and short enough to keep the pipeline tractable. Its absolute value is revisited in the Limitations (Section~\ref{sec:discussion}). Unity's asset-import pipeline recompiles each script up to three times per record. Error lines are deduplicated at the raw-log level before code extraction to avoid triple-counting the same diagnostic.}

\subsection{Worked Example: Stealth}
\label{sec:exemplar}

\begin{figure}[!t]
\centering
\includegraphics[width=0.7\columnwidth]{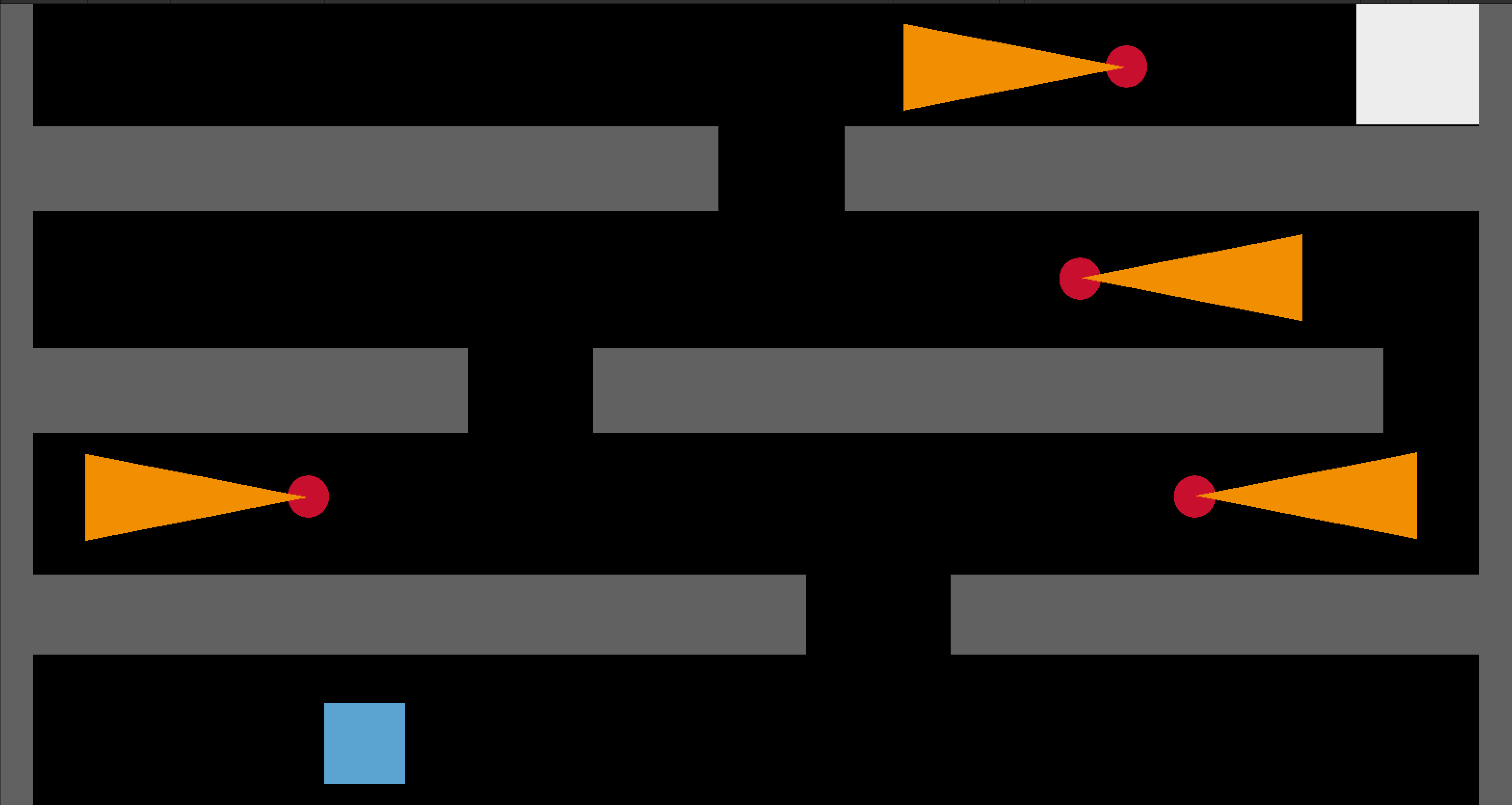}
\caption{Reference Unity instantiation of the Stealth goal pattern. The player (blue square, lower left) must reach the goal region (white square, upper right) while moving within the walled area (grey rectangles as walls) and avoiding the vision cones (orange wedges) of patrolling guards (red circles). Reaching the goal undetected satisfies the pattern's goal structure.}
\label{fig:exemplar}
\end{figure}

We walk through one record from input to failure to illustrate the research design pipeline. The record is Stealth in 30B-Qwen3-Ed (the Editor-style mode) under \texttt{with\_schema\_full}, seed~6. Its input is the \emph{Stealth} pattern description, a multi-paragraph Markdown document that begins ``Stealth is the goal to move through a certain area and perform an action without being detected'' (full text in Appendix~\ref{app:stealth_md}). This document is the only description of the pattern. Figure~\ref{fig:exemplar} shows the reference Unity scene. The model never receives this Markdown on its own, and never the scene, its assets, or their names. The Markdown is filled into a short prompt template (Appendix~\ref{app:prompts}), and generation runs in two calls. Of the five outcomes in Figure~\ref{fig:outcomes}, this record exits at \emph{compile error}. It passes the format sanitizer, reaches compilation, and returns diagnostics rather than timing out. The rest of this section opens that outcome to show what the failure is.

In Step~1 of Section~\ref{subsec:pipeline}, the IR-maker prompt places a short header before the Markdown and asks for an engine-specific Intermediate Representation (IR) as JSON. With the placeholders filled for this record it reads:

{\scriptsize
\begin{verbatim}
[pattern: Stealth]
[method: with_schema_full]

Generate an engine-specific Intermediate
Representation (IR) JSON for the playable
concept described below. Thereafter, you may
refer to it as IR.
Output ONLY valid JSON. No extra text.
<IR schema and full Stealth pattern description>
\end{verbatim}
}
\noindent The model returns the IR below (abridged). Its seven \texttt{scripts} entries name six component classes of its own invention (\texttt{PlayerStealthController}, \texttt{GuardAI}, \dots), each bound to a scene object.

{\scriptsize
\begin{verbatim}
{
"scene": ...,
"objects": [
  {"id":"player_001","name":"Player Character", ...},
  {"id":"guard_001","name":"Guard", ...}, ...
],
"scripts": [
  {"object_id":"player_001",
   "class_name":"PlayerStealthController"},
  {"object_id":"guard_001", "class_name":"GuardAI"}, ...
],
"params": ..., "runtime_params": ...,
"links": ..., "rules": ...
}
\end{verbatim}
}

In Step~2, the coder prompt places the Step-1 IR after its own header and asks for raw C\# only. With the placeholders filled it reads:

{\scriptsize
\begin{verbatim}
[pattern: Stealth]
[method: with_schema_full]

Generate a Unity Editor script that
instantiates a scene matching the following
engine-specific Intermediate Representation
(IR). Thereafter, you may refer to it as IR.
Output only raw C# code.
<full Step-1 IR>
\end{verbatim}
}
\noindent The returned script attaches each component (e.g., \texttt{guard1GO.AddComponent<GuardAI>()}) but never defines any of the six classes, none of which exists in the project. Compiling this script is the attempt to build the scene. It fails with 31 occurrences of \texttt{CS0246} (type or namespace not found), one per reference site. These diagnostics place the record in the \emph{compile error} outcome of Figure~\ref{fig:outcomes}. This is a domain-specific failure. Resolving it requires knowledge of what types the Unity engine and project provide. By contrast, a syntactic error such as a missing semicolon is domain-independent and requires no knowledge of Unity. Section~\ref{subsec:taxonomy} formalises this distinction as the two error classes used throughout, \emph{Grounding} (domain-specific) and \emph{Hygiene} (domain-independent). This record's invented types are a \emph{Grounding} failure.

\section{Results}
\label{sec:results}

This section reports the results of the experiment designed in Section~\ref{sec:design}. We begin with the aggregate failure landscape from record-level outcomes. Each record (defined in Section~\ref{sec:design}) is one concrete run of the experiment, a single generation of C\# script for one [[model, generation mode], condition, pattern, seed]. We categorize the observed error codes as Grounding or Hygiene. Using this Grounding/Hygiene taxonomy as a lens, we read the error-code distribution across the whole corpus, across conditioning levels and [model, generation mode]. We then turn to the per-pattern error structure and bound its range across the whole landscape. A zoom-in on a single [model, generation mode] (30B-Qwen3-Ed) closes the section at finer granularity.

\subsection{The Failure Landscape}

This subsection establishes the Grounding/Hygiene taxonomy that the rest of the paper applies. The record-level outcomes come first and show which records yield compiler evidence at all. The taxonomy is then derived from the 99 observed error codes. The closing distribution and composition views check how the two categories behave across [model, generation mode]. The differences between [model, generation mode] reported along the way are context for reading the per-pattern analysis.

\subsubsection{Record-Level Outcomes}

Table~\ref{tab:outcomes} reports the record-level outcome distribution for all four IR conditioning levels in each [model, generation mode]. No generated C\# script compiled into a runnable scene across any of the 10,400 records. The proportion producing compiler output (\texttt{timeout\_with\_cs}) varies sharply by [model, generation mode], from 18.5\% (771 of 4,160) across 7B-Qwen2.5 and 16B-DeepSeek combined to 45.2\% in 22B-Codestral, 60.7\% in 30B-Qwen3-Ed, and 18.6\% in 30B-Qwen3-Rt.

Table~\ref{tab:outcomes} reveals conditioning responses. For 7B-Qwen2.5 and 16B-DeepSeek, sanitizer rejection climbs with every step of strictness (7B-Qwen2.5: 17.7\%, 70.8\%, 82.5\%, 89.4\%) and is already dominant at \texttt{free}. Any schema pushes most records out of the compiler's reach. 22B-Codestral moves the opposite way, its sanitizer rate \emph{falling} with stricter schemas (64.0\% to 14.4\%) while the share of records with compiler output rises to 74.4\%. Stricter structure improves format compliance even as the generated C\# scripts grow and break. 30B-Qwen3-Ed shows the same direction more mildly, the share with compiler output rising from 23.5\% to 87.7\% while sanitizer rejection stays under 16\%. 30B-Qwen3-Rt collapses under any schema, with sanitizer rejection at 91.9\% by \texttt{free} and 98.1\% at \texttt{full}, leaving almost no records in the error-code pool.

\begin{table}[!t]
\renewcommand{\arraystretch}{1.1}
\caption{Record-level outcomes for every [[model, generation mode], condition], as \% of $n=520$ records per row (rows sum to 100\%). Sanit., sanitizer rejected; Comp.\ err, compile error with captured diagnostics; TO/noCS, timeout, no diagnostics; No entry, compiled but no valid entry point; Pass, compiled with a valid entry point. \textbf{Pass is 0 in every row}, the premise of the study.}
\label{tab:outcomes}
\centering
\scriptsize
\setlength{\tabcolsep}{3pt}
\begin{tabular}{llrrrrr}
\hline
\bfseries [Model, gen.\ mode] & \bfseries Condition &
  \bfseries Sanit. & \bfseries Comp.\ err & \bfseries TO/noCS & \bfseries No entry & \bfseries Pass \\
\hline
7B-Qwen2.5   & no\_schema & 17.7 & 33.8 & 0.0 & 48.5 & 0.0 \\
        & sch\_free  & 70.8 & 21.7 & 0.0 &  7.5 & 0.0 \\
        & sch\_min   & 82.5 & 13.5 & 0.0 &  4.0 & 0.0 \\
        & sch\_full  & 89.4 &  9.8 & 0.0 &  0.8 & 0.0 \\
\hline
16B-DeepSeek & no\_schema &  0.0 & 37.5 & 0.0 & 62.5 & 0.0 \\
        & sch\_free  & 77.5 &  8.5 & 0.0 & 14.0 & 0.0 \\
        & sch\_min   & 84.2 & 11.9 & 0.0 &  3.8 & 0.0 \\
        & sch\_full  & 85.4 & 11.5 & 0.0 &  3.1 & 0.0 \\
\hline
22B-Codestral & no\_schema & 11.5 & 36.0 & 0.4 & 52.1 & 0.0 \\
        & sch\_free  & 64.0 & 13.3 & 0.4 & 22.3 & 0.0 \\
        & sch\_min   & 31.2 & 57.1 & 0.4 & 11.3 & 0.0 \\
        & sch\_full  & 14.4 & 74.4 & 0.4 & 10.8 & 0.0 \\
\hline
30B-Qwen3-Ed & no\_schema &  0.0 & 23.5 & 0.0 & 76.5 & 0.0 \\
        & sch\_free  &  1.0 & 56.9 & 0.6 & 41.5 & 0.0 \\
        & sch\_min   & 15.2 & 74.6 & 0.4 &  9.8 & 0.0 \\
        & sch\_full  &  8.8 & 87.7 & 0.0 &  3.5 & 0.0 \\
\hline
30B-Qwen3-Rt & no\_schema & 12.9 & 67.5 & 0.6 & 19.0 & 0.0 \\
        & sch\_free  & 91.9 &  3.7 & 0.4 &  4.0 & 0.0 \\
        & sch\_min   & 95.6 &  2.7 & 0.0 &  1.7 & 0.0 \\
        & sch\_full  & 98.1 &  0.4 & 0.0 &  1.5 & 0.0 \\
\hline
\end{tabular}
\end{table}

\subsubsection{Observed Error Codes and the Grounding/Hygiene Taxonomy}
\label{subsec:taxonomy}

Across all records with compiler output, 99 distinct C\# error codes appear (each listed with its message template and total count in Appendix~\ref{app:allcodes}). Every code is a standard diagnostic defined by the C\# compiler, with a documented entry in Microsoft's compiler-error reference\footnote{E.g., \texttt{CS0246}: \url{https://learn.microsoft.com/en-us/dotnet/csharp/language-reference/compiler-messages/cs0246}}. Only the Grounding/Hygiene categorization is introduced by this study. They split cleanly into the two categories that define the taxonomy used throughout, \emph{Grounding} and \emph{Hygiene}. The 18 Grounding codes require Unity-specific knowledge to resolve. For example, \texttt{CS0246} (type or namespace not found) names model-invented types such as \texttt{EnemyAI}, \texttt{PlayerScript}, and \texttt{GuardAI} that exist in neither the engine nor the project. \texttt{CS1061} (no such member) names members such as \texttt{health} or \texttt{movementSpeed} that the model assumed a Unity component exposes. The full set spans undeclared types and namespaces (\texttt{CS0246}, \texttt{CS0234}), inaccessible or missing members (\texttt{CS1061}, \texttt{CS0117}, \texttt{CS0122}), incorrect overrides (\texttt{CS0115}, \texttt{CS0534}), and Unity-type mismatches (\texttt{CS0311}, \texttt{CS0315}, \texttt{CS8121}, among others). The remaining 81 Hygiene codes need no Unity knowledge to diagnose. They are structural and syntactic defects such as \texttt{CS1003} (syntax error), \texttt{CS1002} (missing semicolon), unmatched braces, undeclared variables, and invalid expressions. The boundary between the two categories is anchored in the compiler's own diagnostic semantics, not in a post hoc judgement about each record. An error code is Grounding when resolving its diagnostic requires type, member, or namespace information external to the script, which only the Unity engine and project supply. An error code is Hygiene when its diagnostic is purely lexical or syntactic and consults no external type information. Because each code carries a fixed diagnostic meaning, the mapping from code to category is deterministic.
The term \emph{hygiene} is borrowed from programming-language theory~\cite{kohlbecker1986hygienic} and software-engineering practice~\cite{tilbrook1990washing}, where it denotes code that is locally well-formed independent of external domain conventions.

Each error occurrence is one instance of an error code emitted when compiling the generated C\# script. The Grounding/Hygiene category attaches to the error code (a distinct C\# diagnostic such as \texttt{CS0246}), not to the record (one generation attempt for a given [[model, generation mode], condition, pattern, seed]). The two categories are mutually exclusive. Each error code is either Grounding or Hygiene. A single record may still contain both Grounding and Hygiene error codes because compiling one generated C\# script typically produces several distinct errors. Our unit of analysis is the error occurrence, counted with its frequency. For instance, if \texttt{CS0246} appears five times in one record and names a different missing type each time, all five occurrences are counted rather than just one. Per-pattern profiles sum these counts across all 20 seeds within each [[model, generation mode], condition, pattern]. Table~\ref{tab:lineage} traces one pattern from its individual records, through the error occurrences and their codes, to the aggregated per-pattern profile.

\begin{table*}[!t]
\centering
\scriptsize
\caption{Illustrative lineage from records to a per-pattern profile (\emph{Stealth}, \texttt{30B-Qwen3-Ed}, \texttt{with\_schema\_full}). Each occurrence (column~2) instantiates one error code, abstracting the concrete identifier up to the code and its message template (column~3); the Grounding/Hygiene category attaches to the code. The same code \texttt{CS0246} recurs with a different invented type across seeds. Two of the 20 seeds are shown to illustrate the aggregation; the real per-pattern profile sums all 20. Hygiene rows (\texttt{CS1002}, \texttt{CS1003}) are representative.}
\label{tab:lineage}
\begin{tabular}{@{}p{2.4cm} p{2.5cm} p{5.4cm} l p{2.6cm}@{}}
\toprule
\textbf{record} & \textbf{error occurrence} & \textbf{error code and message template} & \textbf{category} & \textbf{per-pattern profile} \\
\midrule
\multirow{2}{=}{$[[$\texttt{30B-Qwen3-Ed}$],$ \texttt{full}, Stealth, $6]$}
 & \texttt{CS0246} $\to$ \texttt{GuardAI} & \texttt{CS0246}: The type or namespace name \textit{X} could not be found & Grounding & \multirow{4}{=}{\emph{Stealth}\\[3pt] \textit{this table (2 seeds):}\\ G 2, H 2 occ\\[3pt] \textit{all 20 seeds:}\\ G 334, H 44 occ\\ (16.7 / 2.2 per seed)} \\[10pt]
 & \texttt{CS1002} $\to$ missing \texttt{;} & \texttt{CS1002}: \texttt{;} expected & Hygiene & \\
\cmidrule(lr){1-4}
\multirow{2}{=}{$[[$\texttt{30B-Qwen3-Ed}$],$ \texttt{full}, Stealth, $7]$}
 & \texttt{CS0246} $\to$ \texttt{EnemyAI} & \texttt{CS0246}: The type or namespace name \textit{X} could not be found & Grounding & \\[10pt]
 & \texttt{CS1003} $\to$ syntax error & \texttt{CS1003}: Syntax error, \textit{X} expected & Hygiene & \\
\bottomrule
\end{tabular}
\end{table*}

\subsubsection{Error Code Distribution}

Of the 90,673 error occurrences, 76\% come from 22B-Codestral alone, whose verbose and structurally broken C\# scripts accumulate far more error lines per compilation attempt than the other models. Table~\ref{tab:top10} lists the top-10 error codes by total occurrence. The most frequent overall is \texttt{CS1003} (syntax error, Hygiene) with 28,571 occurrences, or 31.5\% of the 90,673 total. Of these, 26,268 (91.9\%) come from 22B-Codestral. Next is \texttt{CS1002} (missing semicolon, Hygiene) with 15,943 occurrences (17.6\%). Among Grounding error codes, \texttt{CS0246} (type or namespace not found) dominates with 12,297 occurrences (13.6\%) spread across every [model, generation mode] (505, 279, 3,245, 7,711, and 557 from 7B-Qwen2.5 through 30B-Qwen3-Rt). Its presence in all five marks the Unity type system as a universal knowledge boundary. The second Grounding error code is \texttt{CS1061} (member not found), with 1,590 occurrences (1.8\%). Of these, 1,245 (78.3\%) sit in 30B-Qwen3-Ed. Only \texttt{CS0246} and \texttt{CS1061} of the top 10 are Grounding error codes. The other eight are Hygiene. Hygiene errors accumulate through repetition. The same missing semicolons and unmatched braces recur line after line. Grounding errors are different. Each names a specific missing type and pinpoints a Unity API boundary the model failed to cross.

\begin{table*}[!t]
\renewcommand{\arraystretch}{1.15}
\caption{Top-10 C\# error codes by total occurrence, with abbreviated message templates and a per-[model, generation mode] breakdown. \textit{X} denotes a quoted identifier, shown with a representative example from the logs in parentheses. G\,=\,Grounding; H\,=\,Hygiene. The five [model, generation mode] are defined in Section~\ref{sec:design}.}
\label{tab:top10}
\centering
\scriptsize
\setlength{\tabcolsep}{2pt}
\begin{tabular}{r l p{5cm} c r r r r r r}
\hline
\bfseries Rk & \bfseries Code & \bfseries Message template & \bfseries GH & \bfseries Total &
  \bfseries 7B-Qwen2.5 & \bfseries 16B-DeepSeek & \bfseries 22B-Codestral & \bfseries 30B-Qwen3-Ed & \bfseries 30B-Qwen3-Rt \\
\hline
 1 & CS1003 & Syntax error, \textit{X} (e.g., \texttt{','}) expected & H & 28,571 &   176 &    88 & 26,268 &    62 & 1,977 \\
 2 & CS1002 & ; expected & H & 15,943 &     7 &   312 & 14,674 &   246 &   704 \\
 3 & CS0246 & Type or namespace \textit{X} (e.g., \texttt{GuardGoal}) not found & G & 12,297 &   505 &   279 &  3,245 & 7,711 &   557 \\
 4 & CS1525 & Invalid expression term \textit{X} (e.g., \texttt{''}) & H &  4,524 &     2 &     0 &  4,136 &     6 &   380 \\
 5 & CS1001 & Identifier expected & H &  4,343 &   239 &   110 &  3,776 &    64 &   154 \\
 6 & CS1022 & Type/namespace definition or EOF expected & H &  3,717 &     2 &   125 &  3,424 &     9 &   157 \\
 7 & CS1513 & \} expected & H &  2,852 &     5 &   257 &  2,272 &   294 &    24 \\
 8 & CS1044 & More than one type in a declaration & H &  2,374 &     0 &     0 &  2,047 &     0 &   327 \\
 9 & CS0103 & Name \textit{X} (e.g., \texttt{whatToConceal}) not in context & H &  2,070 &   131 &    22 &     51 & 1,864 &     2 \\
10 & CS1061 & \textit{X} has no member (e.g., \texttt{startPosition}) & G &  1,590 &    37 &    94 &    202 & 1,245 &    12 \\
\hline
\end{tabular}
\end{table*}

\subsubsection{Composition Across Conditioning and [model, generation mode]}
\label{sec:composition}

\begin{figure*}[!t]
\centering
\includegraphics[width=\textwidth]{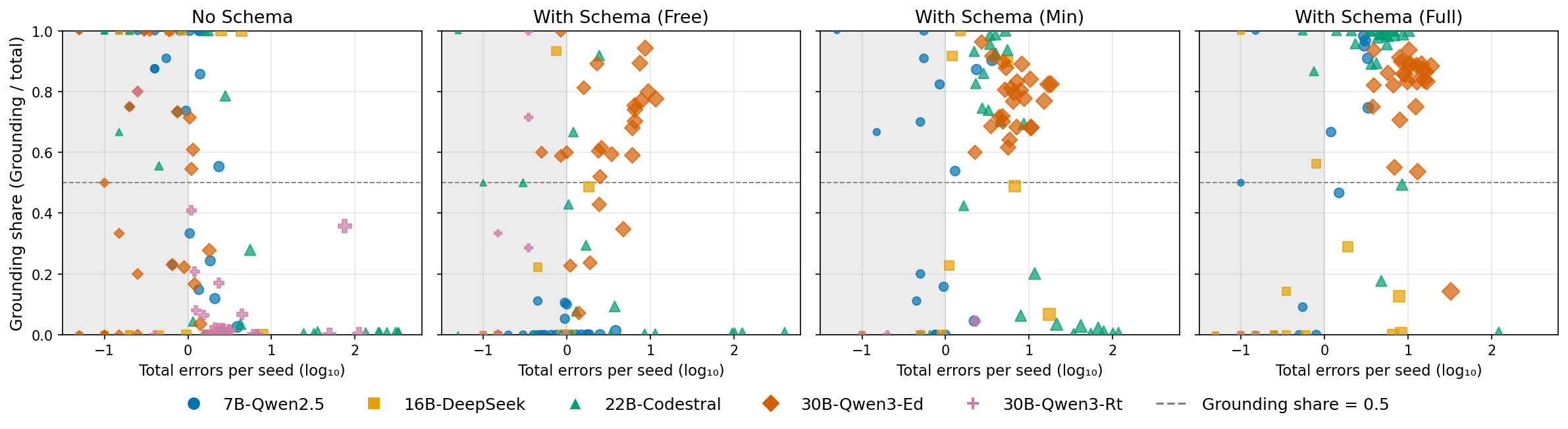}
\caption{Error composition by IR conditioning. Each panel is one conditioning level. Each [[model, generation mode], pattern] appears as a marker, placed by total error occurrences per seed (x, log) and by Grounding share, the fraction of its error occurrences that are Grounding (y). The dashed line at Grounding share~$=0.5$ marks equal Grounding and Hygiene occurrences. Above it, Grounding occurrences are at least half of a marker's total. Markers below one error per seed are shaded. A [[model, generation mode], pattern] with no compiler output is omitted (panel counts 117, 86, 82, and 83).}
\label{fig:scatter}
\end{figure*}

The previous sections ranked individual error codes by volume. From this composition analysis onward, we look only at their Grounding or Hygiene category, not the specific code. Figure~\ref{fig:scatter} traces how the Grounding and Hygiene share (the fraction of error occurrences that are Grounding or Hygiene) of the failures shifts as the schema tightens. Because every error occurrence is either Grounding or Hygiene, the two shares sum to one. The Grounding share alone captures a [[model, generation mode], pattern]'s whole composition. A high Grounding share means failure mostly on missing Unity types and members, the errors that require engine knowledge to resolve. A high Hygiene share means failure on plain syntax, diagnosable without any Unity knowledge. Each panel is one IR condition, holding up to $5\times26=130$ markers, one per [[model, generation mode], pattern] combination. A marker sums the error occurrences over that combination's 20 seed-records, counted per seed. A marker appears only if its records yield at least one error code. A [[model, generation mode], pattern] whose records never produced one is absent, which is why the panel counts fall below 130.

As the schema tightens, fewer markers remain. That means, under a stricter schema, more of a [[model, generation mode], pattern]'s records are rejected at the format sanitizer and never reach the compiler (Section~\ref{subsec:pipeline}). With no records producing an error code, the marker disappears. Among those that remain, fewer reach one error code per seed. This thinning differs sharply by [model, generation mode]. Under \texttt{no\_schema} 117 of the 130 markers are present. 30B-Qwen3-Ed keeps all 26 goal patterns above one error code per seed at \texttt{min} and \texttt{full}. 22B-Codestral keeps 24 to 26. The smaller 7B-Qwen2.5 and 16B-DeepSeek still place 15 and 14 markers at \texttt{full}. Most of those fall below one error code per seed, leaving only 7 and 4 above. 30B-Qwen3-Rt nearly leaves the plot (two markers at \texttt{full}) because almost all its records are sanitizer-rejected. Only 30B-Qwen3-Ed and 22B-Codestral keep full coverage under a strict schema.

Among the markers (i.e., [[model, generation mode], pattern] combination) that remain, composition moves toward Grounding as the schema tightens and converges across the surviving [model, generation mode]. Under \texttt{no\_schema} the markers lean Hygiene. The \emph{Grounding share} quantifies this. No [model, generation mode] reaches a median Grounding share of 0.5 (0.44 for 7B-Qwen2.5, 0.41 for 30B-Qwen3-Ed, near~0 for the other three). These are per-marker medians, taken across a [model, generation mode]'s per-pattern markers, using markers at one error code per seed or more. The clearest trend is 30B-Qwen3-Ed. Its per-marker median Grounding share climbs level by level (0.41, 0.65, 0.79, 0.86). Its markers migrate rightward (median 1.2 to 10.4 error codes per seed) and their spread tightens, forming a small upper-right cluster by \texttt{full}. 22B-Codestral jumps later, from near~0 at \texttt{no\_schema} and \texttt{free} to 0.70 at \texttt{min} and 0.99 at \texttt{full}. Its rise comes through verbose output filled with syntax errors (Section~\ref{sec:discussion}). 7B-Qwen2.5 and 16B-DeepSeek never settle into a clean trend. Strict schema conditioning does not reduce failure. It concentrates the errors that remain onto the Grounding layer.

\subsubsection{The Range of Per-Pattern Error Profiles}

The failure landscape view in the previous sections says nothing about which gameplay goal concepts co-occur with these errors. This section reads the census pattern by pattern. It first bounds the range of per-pattern profiles across all five [model, generation mode], then zooms into 30B-Qwen3-Ed to trace them by pattern, conditioning level, and generation mode.

As in Section~\ref{sec:composition}, we read each error occurrence by its Grounding or Hygiene category, not by its specific error code. The aggregated per-pattern error profile covers five [model, generation mode] $\times$ 26 goal patterns, or 130 [[model, generation mode], pattern] in total. Grounding occurrences exceed Hygiene in 50 of the 130 (38.5\%). Hygiene occurrences exceed Grounding in 79 (60.8\%). One yields no compiler output at all. Across the 130, the Grounding share spans nearly its entire possible range, from 0 to 0.98. At one end a pattern fails purely on plain syntax. At the other it fails almost only on missing Unity references. We examine the two endpoints in detail.

The maximum is 0.98, for Alignment under 22B-Codestral. It has 206 Grounding occurrences (chiefly \texttt{CS0246} with 181 and \texttt{CS1061} with 21) against 5 Hygiene. Both error codes mean a referenced Unity type or member does not exist. With only 5 Hygiene occurrences, the generated C\# scripts are almost free of structural defects. They fail at their Unity references instead, the same invented-vocabulary failure as in the worked example (Section~\ref{sec:exemplar}). 22B-Codestral inflates its raw totals elsewhere with verbose and repetitive output (Section~\ref{sec:discussion}). Repetition can inflate a Grounding count. It cannot produce the near-absence of Hygiene occurrences. The imbalance here is a property of composition, not volume.

The minimum is 0, for Delivery under 16B-DeepSeek. It has 0 Grounding against 37 Hygiene, split between \texttt{CS0101} (19, duplicate type in a namespace) and \texttt{CS0111} (18, duplicate member in a type). The model re-declares the same types and members until the compiler rejects the file. The scripts collapse at the structural layer before any Unity reference is attempted.

The two endpoints differ in model as well as in pattern. The maximum arises under 22B-Codestral and the minimum under 16B-DeepSeek. Between the two models, size, family, and pretraining corpus all change (Section~\ref{sec:design}). The range mixes pattern effects with [model, generation mode] effects.

\subsection{A Zoom-in: 30B-Qwen3-Ed}

To read the pattern dimension cleanly, this section uses \textbf{30B-Qwen3-Ed} alone, for three reasons. First, it carries the cleanest signal. Its sanitizer rejection stays below 16\% across all four IR conditioning levels (Table~\ref{tab:outcomes}). At least 84\% of its records clear the format gate at every level, so its per-pattern readings are the least affected by the compiler-reach selection discussed in the Limitations (Section~\ref{sec:discussion}). The other four lose 85--98\% of records to the format gate under the stricter schemas (\texttt{with\_schema\_min} and \texttt{with\_schema\_full}), or are swamped by verbose output that inflates their counts (Section~\ref{sec:discussion}). Second, it has the broadest coverage, populating 101 of the $26\times4$ pattern-by-condition combinations, more than any other [model, generation mode]. Third, it is the largest model in the study. Its errors show how difficulty distributes across patterns once the model has its best chance, the most diagnostic setting for isolating the pattern dimension (Section~\ref{sec:design}). The zoom-in reads this single [model, generation mode] from three views: by pattern, across the IR conditioning levels, and across the two generation modes.

The first view is by pattern. In 30B-Qwen3-Ed, the highest per-seed counts of Grounding error occurrences fall on Rescue (9.01), Exploration (8.11), Survive (7.59), and Stealth (6.69). Capture is the contrasting case, with an elevated per-seed Hygiene count (7.88) against a comparatively low per-seed Grounding count (2.38). For the four patterns with the highest per-seed Grounding counts, the most frequent error code is \texttt{CS0246} (type or namespace not found). The identifiers it names extend the invented vocabulary of the worked example (Section~\ref{sec:exemplar}). Each is named after its pattern's own mechanics: \texttt{GuardAI} and \texttt{AlarmSystem} for Stealth, \texttt{RescueTarget} for Rescue, \texttt{ExplorationTracker} for Exploration, and \texttt{HealthSystem} for Survive. The names encode the correct design semantics. None of these types exists in the engine or the project. The Grounding boundary is referential, not conceptual. It is a gap between what the pattern requires and what the model knows the engine to provide. The record-level outcomes vary less across these patterns than the code content does. Most reach the compiler at similar rates (50--58 of 80 records yield compiler diagnostics), while Capture and Ownership show higher sanitizer rejection (8 and 10 of 80) and correspondingly thinner code evidence.

\begin{figure}[!t]
\centering
\includegraphics[width=0.9\columnwidth]{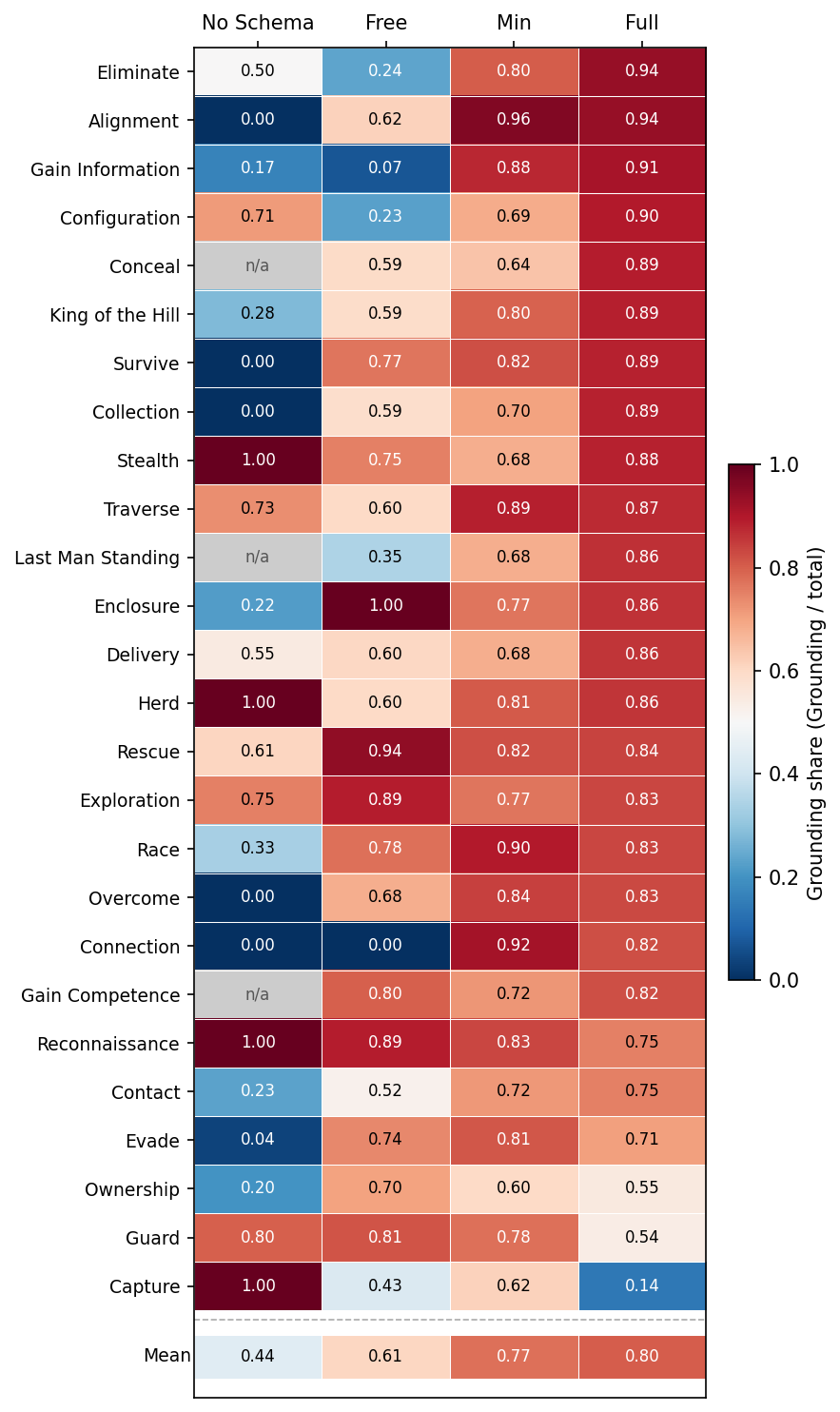}
\caption{Per-pattern error composition across IR conditioning levels, read from 30B-Qwen3-Ed. Each row is a goal pattern and each column an IR conditioning level. Each [model, generation mode] pools the error occurrences of all seeds for that pattern and level. The shade is the [model, generation mode]'s Grounding share (Section~\ref{sec:composition}): blue, Grounding share $<$ 0.5; red, Grounding share $>$ 0.5; white, the 0.5 balance. Rows are ordered by \texttt{full} Grounding share. Three \texttt{no\_schema} entries with no compiler output are marked n/a. The bottom row gives the column mean.
}
\label{fig:heatmap}
\end{figure}

The second view follows the IR conditioning axis. Figure~\ref{fig:heatmap} reads 30B-Qwen3-Ed pattern by pattern. The shift toward Grounding appears again, now within a single [model, generation mode]. The column-mean Grounding share rises from 0.44 (\texttt{no\_schema}) through 0.61 (\texttt{free}) and 0.77 (\texttt{min}) to 0.80 (\texttt{full}). Each column mean weights a pattern by its error occurrences, so its values differ from the equally weighted per-marker medians read off Figure~\ref{fig:scatter}. Most of the rise comes between \texttt{free} and \texttt{min}. After \texttt{min} the share levels off. Of the 23 patterns with data at both extremes, 18 end higher in Grounding share under \texttt{full} than under \texttt{no\_schema}. The schema scaffolds the syntactic layer. Failures that would otherwise surface as Hygiene defects are displaced upward to the Grounding layer, where the model's Unity knowledge becomes the limiting factor.

The shift is a tendency, not a law. Five patterns (Stealth, Herd, Reconnaissance, Guard, and Capture) move the opposite way, ending \emph{lower} in Grounding share under \texttt{full}. Capture is the sharpest case, falling from a Grounding share of 1.00 under \texttt{no\_schema} to 0.14 under \texttt{full}. The error codes explain the reversal. Under \texttt{no\_schema}, Capture produces only 7 error occurrences in total (5 \texttt{CS0122}, inaccessible member; 2 \texttt{CS0246}), all Grounding. The generated C\# scripts are short and almost correct, failing only on a few Unity references. Under \texttt{full}, the same pattern produces 652 occurrences, 559 of them Hygiene, mostly unmatched braces (\texttt{CS1513}, 235) and missing semicolons (\texttt{CS1002}, 223). The schema drives the model to emit substantially longer and more structured C\# scripts that collapse at the syntactic layer before any Unity reference is reached. The mechanism is the same one that moves most patterns upward (schema conditioning raises code complexity). Its direction depends on whether the model can keep the larger output syntactically intact for the pattern at hand.

The third view switches the generation mode. It compares Qwen3-Coder-30B under its two modes, Editor-style (30B-Qwen3-Ed) and Runtime-builder (30B-Qwen3-Rt), holding condition, pattern, and seed constant. 30B-Qwen3-Ed produces 12,160 error occurrences across 1,262 records with compiler output. 30B-Qwen3-Rt produces 5,989 occurrences across 386 such records. The lower Runtime figures reflect the sharp increase in sanitizer rejection under schema conditioning (Table~\ref{tab:outcomes}). At the per-pattern level (Table~\ref{tab:f1}), the high per-seed Grounding counts noted above (Rescue, Exploration, Survive, Stealth) belong to the Editor mode. The Runtime-builder mode substantially reduces per-seed Grounding counts for most patterns but introduces severe Hygiene spikes for several patterns: Collection (per-seed Hygiene count 28.61 vs.\ 0.85 in Editor) and Connection (12.40 vs.\ 0.44). Configuration rises on both (Grounding 6.73 vs.\ 2.73, Hygiene 12.11 vs.\ 0.78). These Hygiene spikes mark structurally broken generated C\# scripts, with no accompanying improvement in compilation outcomes.

\begin{table}[!t]
\renewcommand{\arraystretch}{1.1}
\caption{Per-pattern comparison of the two generation modes (30B, all IR levels combined). Ed-G / Ed-H: per-seed counts of Grounding and Hygiene error occurrences in the Editor mode (30B-Qwen3-Ed). Rt-G / Rt-H: the same in the Runtime-builder mode (30B-Qwen3-Rt). Per-seed counts rounded to two decimal places.}
\label{tab:f1}
\centering
\scriptsize
\begin{tabular}{lrrrr}
\hline
\bfseries Pattern & \bfseries Ed-G & \bfseries Ed-H & \bfseries Rt-G & \bfseries Rt-H \\
\hline
Ownership            &  2.45 &  1.54 &  0.01 &  0.64 \\
Collection           &  3.78 &  0.85 &  0.09 & 28.61 \\
Eliminate            &  4.13 &  0.93 &  0.00 &  0.76 \\
Capture              &  2.38 &  7.88 &  0.00 &  0.10 \\
Overcome             &  5.88 &  1.46 &  0.01 &  0.78 \\
Evade                &  3.88 &  1.64 &  0.00 &  0.65 \\
Stealth              &  6.69 &  1.51 &  0.00 &  0.48 \\
Herd        &  3.20 &  0.69 &  0.05 &  0.01 \\
Conceal              &  3.81 &  1.16 &  0.11 &  0.16 \\
Rescue              &  9.01 &  1.63 &  0.10 &  0.49 \\
Delivery            &  4.19 &  1.51 &  0.00 &  0.58 \\
Guard               &  3.83 &  2.09 &  0.08 &  1.04 \\
Race                &  5.41 &  1.21 &  0.01 &  0.58 \\
Alignment      &  1.96 &  0.35 &  0.06 &  1.74 \\
Configuration       &  2.73 &  0.78 &  6.73 & 12.11 \\
Traverse            &  5.38 &  0.81 &  0.00 &  0.40 \\
Survive             &  7.59 &  1.61 &  0.03 &  0.79 \\
Connection    &  2.20 &  0.44 &  0.00 & 12.40 \\
Exploration         &  8.11 &  1.79 &  0.00 &  0.68 \\
Reconnaissance      &  4.48 &  1.14 &  0.11 &  0.29 \\
Contact             &  1.86 &  0.98 &  0.03 &  0.39 \\
Enclosure           &  2.74 &  0.75 &  0.01 &  0.51 \\
Gain Competence      &  3.55 &  0.99 &  0.06 &  0.28 \\
Gain Information     &  3.05 &  0.91 &  0.01 &  0.89 \\
Last Man Standing     &  5.71 &  2.16 &  0.00 &  0.45 \\
King of the Hill       &  5.56 &  1.69 &  0.03 &  1.56 \\
\hline
\end{tabular}
\end{table}

Overall, switching to the Runtime-builder mode does not eliminate Grounding errors. It redistributes the error landscape. Where the errors land depends on the pattern. For Collection and Connection, the Runtime-builder mode replaces Grounding errors with large Hygiene counts. For Configuration, both error categories are amplified.

\section{Discussion}
\label{sec:discussion}

\subsection{Interpretation of Per-Pattern Profiles}

The per-pattern Grounding share is a diagnostic lens on the Unity-specific knowledge each goal pattern demands. Its two ends fail at different layers. The four patterns with the highest per-seed Grounding error occurrence counts (Rescue, Exploration, Survive, Stealth) share one failure mode. The model posits a dedicated engine-specific component vocabulary and references it as though Unity provided it, producing \texttt{CS0246}. The reference implementations realize the same goals far more simply, with 2D trigger colliders and basic movement. The model understands the concept well enough to reference the \emph{right kind} of operation but fails to name a class or method that exists, a knowledge-boundary failure, not a structural one. The invented names are not copied out of the pattern descriptions. The Rescue description (``free someone or something that is guarded'', Table~\ref{tab:patterns}) nowhere mentions pathfinding or vision. The model still invents \texttt{Pathfinding} and \texttt{ComputerVisionSystem} for Rescue and attaches them as components. The Survive description contains neither collision nor health, yet the model invents \texttt{AvatarCollisionHandler} and \texttt{AvatarHealth}. None of these types exists in the engine or the project. The model infers from the pattern's goal what gameplay machinery a scene would need, then invents engine types to supply it. That inference is itself a grounding failure, a claim about what the engine provides rather than an echo of the prompt. Patterns with a low Grounding share, such as Capture, invert this. Their generated C\# scripts collapse structurally before reaching the layer where Unity knowledge would be tested, as their mechanics reduce to general object-state tracking rather than engine-specific APIs.

The gameplay pattern design language itself supports this reading at the two poles. Bj\"ork and Holopainen's catalogue lists the patterns each goal pattern instantiates and is modulated by~\cite{bjork2005patterns}. These relations spell out what a playable form of the pattern must simulate. Stealth (``move through an area and act without being detected'', Table~\ref{tab:patterns}) instantiates \emph{Movement} (``the action of moving game elements in the Game World''), \emph{Area Control} (``being in control over who can move within an area in the game world''), and \emph{Tension} (``the feeling of caring about the outcome of actions or events in a game without having full control over them''). The catalogue further describes it as a compound goal pattern built from \emph{Conceal} (``hinder other players' ability to gain information'') and \emph{Evade} (``avoid being captured or hit''). Therefore, a scene that realizes Stealth must simulate concealment, evasion, and detection in some form. The relations do not dictate one embodiment. Read literally in a 2D scene, as our reference instantiation does, they become several mechanics at once: moving guards, fields of vision, and a rule for detection. In Unity that machinery lives in the perception and physics APIs, which is exactly where the model invents types such as \texttt{GuardAI} and \texttt{AlarmSystem}. Capture (``eliminate or take ownership of an actively resisting goal object'') instantiates \emph{Gain Ownership} (``gain ownership of a game element''), \emph{Transfer of Control} (``when the influence over a game element is passed from one player to another''), and \emph{Combat} (``actions where the intent is to kill or otherwise overcome opponents''), relations that revolve around who owns or controls a game element. A scene that realizes Capture can keep that state in plain C\# variables. Little of the engine is needed. Therefore, little of it can be named wrongly. Our per-pattern reading of the Grounding/Hygiene census turns the goal-pattern vocabulary of Bj\"ork and Holopainen~\cite{bjork2005patterns}, realized here as GPCs~\cite{lyu_gpc_2023}, into an empirical ordering of design concepts by the engine knowledge each demands.

\subsection{Model Scale and Identity}

From 7B to 30B, no model produces a runnable scene in a single pass. Within the deployable open-weight tier, a larger model is not a remedy. The trend is familiar from code-generation benchmarks. Pass rates climb with model size on self-contained routines~\cite{chen2021evaluating,austin2021program}. However, accuracy degrades sharply once the model must name real types and signatures from an external API~\cite{jiang2026survey,jimenez2023swe}. Engine-coupled generation, which constructs a live scene, sits at the far end of that axis. What varies by [model, generation mode] is the \emph{composition} of failure. It does not track parameter count. 22B-Codestral, larger than 7B-Qwen2.5 and 16B-DeepSeek, fails almost entirely at the syntactic layer. 30B-Qwen3-Ed reaches the Grounding layer. The two mixture-of-experts models differ again (Section~\ref{sec:results}). The sharpest case is at fixed scale. The same Qwen3-Coder-30B model reaches the compiler in the Editor-style mode but collapses to near-total sanitizer rejection in the Runtime-builder mode. These differences are driven by model identity and generation mode, not scale.

One model-specific behavior distorts count-based readings. 22B-Codestral contributes 68,813 of the 90,673 occurrences (76\%) from only 2,080 records, with single records emitting up to 2,066 diagnostics. In its long records the model stops writing C\# partway and continues in the style of the prompt's documentation, emitting markdown that the C\# compiler rejects line by line. This is an artifact of long output, not denser Unity API errors. We report per-seed counts of error occurrences and the Grounding share rather than raw totals, a caution for count-based profiling of LLM-generated code generally~\cite{tambon2024bugs,dou2024whats}.

For a team choosing a model, the takeaway is concrete. Within the studied tier, a larger model buys a different failure composition, not a runnable scene. Under a single-pass constraint, choosing a model is choosing which failure layer to face. A model that fails at the syntactic layer calls for output repair. A model that fails at the Grounding layer calls for engine knowledge that has to come from outside the model. The generation-mode contrast is measured only at 30B, where the Editor mode reaches the compiler far more often (1,262 against 386 records with compiler output). Smaller models ran in the Editor mode only, leaving the mode question below 30B open. Below the zero-compilation floor, pass-rate evaluation stops discriminating. The error composition is the signal that remains. This is what a failure census adds to pass-rate benchmarks of LLM code generation, and what it offers PCG evaluation at the game-design layer of goals~\cite{hendrikx2013procedural}: a map of where generator capability ends before any content exists to evaluate.

\subsection{What IR Conditioning Buys, and What It Costs}

Like model scale and the generation mode in the previous section, IR conditioning never turns failure into success. In every [model, generation mode], adding a schema changes only \emph{where} a record fails, not whether it fails. We use the IR to expose a different layer of failure for measurement.

What the schema changes depends on the model. The two smallest, 7B-Qwen2.5 and 16B-DeepSeek, mostly stop producing usable output at all. Any schema sends most of their records to sanitizer rejection (70.8\% and 77.5\%). Under the strictest schema, most of those are not broken scripts but text that has stopped being code (a single token repeated, or a lone code-fence marker). Whatever Unity knowledge these models have, the schema makes it impossible to observe. 22B-Codestral and 30B-Qwen3-Ed do the opposite. A stricter schema helps them clear the format gate (22B-Codestral's sanitizer rate drops from 64.0\% to 14.4\%). More records reach the compiler. They fail there too, with richer diagnostics but no runnable scene. The same Qwen3-Coder-30B in the Runtime-builder mode collapses to 91.9--98.1\% sanitizer rejection. Even for one fixed model, the schema's effect depends on the generation mode. For the records that do reach the compiler, the failures move from syntax toward Unity knowledge as the schema tightens (mean Grounding share 0.44 to 0.80, Figure~\ref{fig:heatmap}). The schema cleans up the syntax and leaves the errors that need real engine knowledge, relocating the failure rather than removing it.

This is the limit of structural conditioning. A schema constrains the shape of the output but supplies no Unity knowledge. It can move failure to the Grounding layer but cannot close the Grounding gap that blocks compilation. An IR that actually improved compilation would have to carry engine facts (real types, signatures, and members), not just structure.

The direction of the schema's effect appears to depend on model capacity. The schema is a structural contract: required fields, names, and shapes, with no engine content. Holding to that contract while still writing C\# is itself a demand on the model. For 22B-Codestral and 30B-Qwen3-Ed, the contract keeps the output in code form, which lets more records survive the format gate. For 7B-Qwen2.5 and 16B-DeepSeek, the same contract is more than the model can hold. Their output falls apart before any C\# appears. The same dependence is reported elsewhere. Format restrictions degrade generation, with stricter restrictions degrading it more~\cite{tam2024speakfreely}. Structured-output reliability collapses as schemas grow more complex~\cite{geng2025jsonschemabench}. In this study the threshold sits between 16B and 22B. Having a contract matters more than how deep it goes. Most of the composition shift arrives by \texttt{min} (Figure~\ref{fig:heatmap}). The Capture reversal (Section~\ref{sec:results}) marks where added depth turns against the model.

\subsection{Implications for LLM-Assisted Game Scene Development}

The Grounding/Hygiene taxonomy is a practical diagnostic for choosing interventions. Patterns with a high Grounding share, where the primary error is missing or incorrect Unity API references, are unlikely to benefit from generic prompt engineering or syntactic post-processing. They may require engine-knowledge augmentation such as retrieval over Unity documentation~\cite{zhou2023docprompting}, fine-tuning on Unity corpora, or structured knowledge injection~\cite{jiang2024selfplanning}. The 30B Editor-versus-Runtime comparison illustrates this. Switching the generation mode redistributes rather than eliminates Grounding errors, and for several patterns (Collection, Connection) introduces severe Hygiene spikes without fixing the missing engine knowledge underneath. In gameplay terms, patterns with a high Grounding share are those whose core mechanic is coupled to the engine itself. Stealth's detection, Exploration's spatial search, and Survive's collision and health all have to be expressed through specific Unity perception and physics components, exactly where the model invents nonexistent types. Meanwhile, Capture keeps a low Grounding share because its mechanic reduces to tracking who owns what in plain variables, with no such coupling. A designer can anticipate where single-pass generation will break by asking whether a concept's defining mechanic lives in the engine's perception and physics layer or in plain state logic. They can then budget engine-knowledge support (or hand-authoring) for the former.

Patterns with a low Grounding share, by contrast, are better candidates for constrained or grammar-guided decoding~\cite{wang2023grammar}, output sanitization, or syntactic post-processing. Their errors do not require Unity knowledge to diagnose or correct. A rule-based sanitizer on general C\# syntax can in principle intercept most Hygiene errors.

The record-level outcomes add a deployment caution. On this evidence, pairing strict schema conditioning with sub-20B models is wasted computation. Of such records, 85--90\% are lost at the format gate before any script compiles (Table~\ref{tab:outcomes}). Teams constrained to small open-weight models should either drop schema conditioning or invest first in format compliance (e.g., grammar-constrained decoding) before structured knowledge injection can pay off.

The single-pass results also show what an iterative repair loop would have to add. A pattern with a high Grounding share forces each repair loop to supply Unity API knowledge the model cannot supply itself. A pattern with a low Grounding share is easier to repair because the model already has enough Unity knowledge for a locally targeted fix. The contemporaneous Mage benchmark, built on a related class of goal-pattern tasks with its own scenes and pipeline, similarly finds near-zero compilation rates~\cite{mage2026anon}. The floor is not an artifact of our particular instantiations or harness. Whether it holds beyond goal-pattern tasks in Unity is untested.

\subsection{Limitations}

Several limitations bound these findings. \textbf{(a) Pattern instantiation vs.\ pattern.} Each goal pattern is an abstract design construct. Our 26 targets are its current Unity instantiations, one per pattern, so the error profiles characterize \emph{these particular instantiations}, not goal patterns in general. All 26 are 2D scenes built from the same engine primitives, compiled through the same harness, and prompted through the same template. A cross-pattern difference in the census is unlikely to come from one pattern receiving a more elaborate implementation. This control is at the implementation level and does not make a single instantiation representative of its pattern. \textbf{(b) Model axis is not a controlled scale sweep.} Size, family, pretraining corpus, and architecture (the 16B and 30B are sparse mixture-of-experts, with 2.4B and 3.3B active parameters) change together along the axis. Differences by [model, generation mode] cannot be attributed to parameter count alone. The uniform compilation failure is unaffected, as every model fails regardless of family or size. Claims that separate scale from model identity are out of scope. \textbf{(c) Census coverage and compiler-reach selection.} The census covers only records that reach the compiler and emit diagnostics. The share of records that do varies sharply across the [model, generation mode] and conditioning grid. No-entry records compile but expose no entry point for the BatchRunner, typically a plausible gameplay \emph{component} rather than a scene builder (e.g., a well-formed \texttt{public class Contact : MonoBehaviour} that never builds the scene). This is a prompt-specification failure, reported in Table~\ref{tab:outcomes} but contributing no error codes. Sanitizer rejection further removes 70--98\% of records in several [[model, generation mode], condition] rows. A per-pattern Grounding share is read as conditional on producing compiler output, not as an unconditional property of the pattern. This selection is distinct from survivorship bias, which the all-fail corpus removes (Section~\ref{sec:intro}). It acts earlier, on which failed records yield compiler evidence. The reference choice of the zoom-in, with sanitizer rejection below 16\%, keeps it minimal (Section~\ref{sec:results}). \textbf{(d) IR attribution and schema quality.} Under schema conditioning the same LLM generates the IR (Step~1) and the C\# script (Step~2), so compiler evidence alone cannot attribute a Grounding error to a hallucinated IR type or to the coding step. Settling that attribution would require running Step~2 with ground-truth IRs. The results also characterize one frozen schema (v0.2-runtime-evidence), built from the reference implementations and frozen before any run, which guards against tuning to outcomes but does not make it the best possible schema. The headline result does not rest on the schema. \texttt{no\_schema} supplies no IR and still yields zero runnable scenes. \textbf{(e) Timeout budget.} The 120\,s compilation budget is uniform across conditions, which protects comparisons. Its absolute value remains a choice that a replication could probe. \textbf{(f) Single-pass ceiling.} The Grounding/Hygiene profiles characterize the \emph{intrinsic capability ceiling} before any intervention, not performance in real workflows with iterative repair. \textbf{(g) Reading the Grounding share.} A high Grounding share does not mean a model handles Grounding well, only that the syntax is clean enough for Grounding errors to be what is left. A low share usually means the generated C\# scripts broke before they ever referenced Unity. The share marks \emph{where} a [model, generation mode] fails, not how well it does anything. The same reading limits the schema's help. For Capture, the structure the schema asks for is more than the model can write cleanly. Its Grounding share falls from 1.00 to 0.14 as syntax errors return (Section~\ref{sec:results}).

\section{Conclusion}
\label{sec:conclusion}

We conducted a controlled single-pass evaluation of Unity C\# scene generation across 10,400 records (five [model, generation mode]; four IR conditioning levels; 26 goal patterns; 20 seeds). No generated C\# script compiled into a runnable scene. From the failed compilations we extracted 90,673 error occurrences across 99 error codes (18~Grounding, 81~Hygiene). The per-pattern census orders Bj\"ork and Holopainen's goal patterns by the engine knowledge their current specific Unity instantiations demand. Patterns coupled to Unity's perception and physics APIs (e.g., \emph{Stealth}, \emph{Rescue}) concentrate Grounding errors. Patterns reducible to plain state manipulation (e.g., \emph{Capture}) concentrate Hygiene errors. This ordering is an interpretation of the error evidence, not a direct measurement of pattern complexity. A larger model, a stricter schema, and a different generation mode each shift the error profiles without producing a compiling scene. IR conditioning supplies structure rather than engine knowledge. The bottleneck is the missing engine knowledge, not structure, generation mode, or scale. The resulting Grounding/Hygiene error taxonomy gives designers a concrete read on which gameplay concepts current LLMs can realize and which demand engine-knowledge support.

Future work will extend the evaluation in three directions. First, multi-turn repair, feeding compiler errors back to the model, will test how many rounds it takes to reach compilation and whether the Grounding/Hygiene profile predicts which patterns are easy to repair. Second, running the coding step (Step~2) on ground-truth IRs will separate IR-generation failures from code-generation failures. Third, expanding to more patterns and to larger or closed-source models will test whether the findings hold beyond the 7B--30B open-weight tier. Beyond automated repair, the census also serves human-in-the-loop workflows. A person audits and approves the generated C\# scripts, keeping accountability with people, not with the model.

\bibliographystyle{IEEEtran}
\bibliography{refs}

\section{Acknowledgement}

The batch compilation pipeline adapts the write-to-asset approach introduced in AICommand by Keijiro Takahashi~\footnote{\url{https://github.com/keijiro/AICommand}}.

We thank Staffan Björk and Jussi Holopainen for their input on goal playable concepts and related background.

This work was partially supported by the Wallenberg AI, Autonomous Systems and Software Program – Humanities and Society (WASP-HS) funded by the Marianne and Marcus Wallenberg Foundation and the Marcus and Amalia Wallenberg Foundation.

The computations and data handling were enabled by resources provided by the National Academic Infrastructure for Supercomputing in Sweden (NAISS), partially funded by the Swedish Research Council through grant agreement no. 2022-06725.

\appendices

\section{IR Schema: Iteration History and Definition}
\label{app:ir}

Table~\ref{tab:ir_iterations} records the five-iteration history from the initial static draft to the frozen v0.2-runtime-evidence schema. The three \texttt{with\_schema} conditions expose progressively more of this frozen schema to the IR-generation step. \texttt{free} supplies no schema text, so the model structures the IR itself; \texttt{min} supplies only the top-level field names below; and \texttt{full} supplies the full field definitions and hard constraints below.

\begin{table*}[t]
\centering
\caption{IR schema iteration history from initial draft to frozen v0.2-runtime-evidence.}
\label{tab:ir_iterations}
\small
\begin{tabularx}{\linewidth}{@{}c p{0.8cm} X X X@{}}
\toprule
\# & Iteration & What Changed & Why & Impact \\
\midrule
1 & \textbf{v0 static draft} &
Defined six top-level fields: \texttt{scene}, \texttt{objects}, \texttt{scripts}, \texttt{params}, \texttt{links}, \texttt{rules}. &
Minimal structured representation between pattern description and Unity code generation. &
Established baseline schema consumed by all pipeline stages. \\
\addlinespace
2 & \textbf{MVP narrowing} &
Deferred \texttt{params} extraction; pipeline operates on \texttt{objects}, \texttt{scripts}, \texttt{links}, \texttt{rules} only. \texttt{params} emitted as \texttt{\{\}}. &
\texttt{params} requires GUID resolution (scene $\to$ prefab $\to$ script $\to$ \texttt{.cs}), blocking the initial pipeline. &
Unblocked end-to-end generation without serialized-field extraction. \\
\addlinespace
3 & \textbf{Runtime extension} &
Added \texttt{PrefabInstance}/\texttt{PrefabAsset} object types, script-defined \texttt{rules}, and \texttt{runtime\_params} field. &
Static \texttt{.unity} parsing misses prefab-driven gameplay; core behavior emerges from prefab instantiation and runtime script logic. &
IR captures the actual gameplay loop; generation can reason about spawned entities and runtime configuration. \\
\addlinespace
4 & \textbf{Per-instance constraint} &
\texttt{scripts[].object\_id} must reference a real \texttt{objects[].id}. No implicit aggregate placeholders (e.g.\ \texttt{circle\_all}). &
Aggregate placeholders create ambiguous references unresolvable during code generation or evaluation. &
Enforces 1:1 script-to-object binding; enables automatic referential integrity validation. \\
\addlinespace
5 & \textbf{Evidence-aware semantics} &
Conditional relation labels (e.g.\ \texttt{can\_trigger\_game\_win\_if\_aligned}). Required \texttt{evidence\_type} on every \texttt{rules[]} entry. Optional \texttt{confidence} field. &
Unconditional labels over-assert determinism for conditional code paths. Evidence attribution improves trust calibration in generated output. &
Generation produces more accurate causal claims; evaluation can filter or weight rules by evidence type and confidence. \\
\bottomrule
\end{tabularx}
\end{table*}

\textbf{Top-level fields (all required).}
\begin{verbatim}
scene: string
objects: [{ id, name, type }, ...]
scripts: [{ id, object_id, class_name }, ...]
params: {}
runtime_params: { "<script_id>": { ... }, ... }
links: [{ source, target, relation,
          evidence_type? }, ...]
rules: [{ id, type, description, pattern,
          evidence_type, confidence? }, ...]
\end{verbatim}

\textbf{Hard constraints.}
\begin{verbatim}
1) scripts[].object_id must reference
   objects[].id
2) scripts are per-instance
   (no sharing across objects)
3) no implicit aggregate placeholders
4) rules[].evidence_type is required,
   in { direct_code, scene_override, inferred }
\end{verbatim}

\section{Full Error Code Census}
\label{app:allcodes}
\onecolumn
\begingroup\footnotesize
\setlength{\LTcapwidth}{\textwidth}
\begin{longtable}{@{}l r p{0.70\textwidth}@{}}
\caption{All 99 observed C\# compiler error codes, categorized as Grounding (G) or Hygiene (H), with total occurrence count and message template. Quoted identifiers are shown as \textit{X}; for Grounding codes, a representative identifier from the logs is given in parentheses, almost all of them model-invented, pattern-named types absent from the engine and project.}\label{tab:allcodes}\\
\toprule \bfseries Code & \bfseries Total & \bfseries Message template \\ \midrule
\endfirsthead
\multicolumn{3}{@{}l}{\footnotesize\itshape Table~\ref{tab:allcodes} continued from previous page}\\
\toprule \bfseries Code & \bfseries Total & \bfseries Message template \\ \midrule
\endhead
\midrule \multicolumn{3}{r@{}}{\footnotesize\itshape continued on next page}\\ \endfoot
\bottomrule \endlastfoot
\multicolumn{3}{@{}l}{\textit{Grounding (G), 18 codes}}\\
\texttt{CS0246} & 12,297 & The type or namespace name \textit{X} (e.g., \texttt{EnemyAI}, \texttt{GuardAI}) could not be found \\
\texttt{CS1061} & 1,590 & \textit{X} (e.g., \texttt{PlayerController.health}) does not contain a definition for \textit{X} and no accessible extension m \\
\texttt{CS0115} & 213 & \textit{X} (e.g., \texttt{StealthGoal.IsCompleted()}): no suitable method found to override \\
\texttt{CS0234} & 138 & The type or namespace name \textit{X} (e.g., namespace \texttt{Game}) does not exist in the namespace 'UnityEngine.Animatio \\
\texttt{CS0117} & 71 & \textit{X} (e.g., \texttt{RuntimeSceneBuilder}) does not contain a definition for \textit{X} \\
\texttt{CS0122} & 68 & \textit{X} (e.g., \texttt{RescueGoal.Rescue()}) is inaccessible due to its protection level \\
\texttt{CS0311} & 66 & The type \textit{X} cannot be used as type parameter \textit{X} in the generic type or \\
\texttt{CS0030} & 20 & Cannot convert type \textit{X} (e.g., \texttt{DetectableElement[]}) to \textit{X} \\
\texttt{CS0619} & 13 & \textit{X} (e.g., \texttt{GameObject.light}) is obsolete: 'GameObject.AddComponent with string argume \\
\texttt{CS0239} & 8 & \textit{X}: cannot override inherite \\
\texttt{CS0315} & 7 & The type \textit{X} cannot be used as type parameter \textit{X} in the generic type or me \\
\texttt{CS8121} & 4 & An expression of type \textit{X} cannot be handled by a pattern of type 'ScriptableObject \\
\texttt{CS0506} & 3 & \textit{X} (e.g., \texttt{PlayerController.Update()}): cannot override inherited member 'PlayerController.Upd \\
\texttt{CS0534} & 3 & \textit{X} (e.g., \texttt{ConfigurationRule.IsSatisfied()}) does not implement inherited abstract member 'ConfigurationRule.IsSatisfied(Li \\
\texttt{CS1624} & 2 & The body of \textit{X} (e.g., \texttt{TraverseInstantiator.MoveGameElement()}) cannot be an itera \\
\texttt{CS0426} & 1 & The type name \textit{X} does not exist in the type \textit{X} \\
\texttt{CS0509} & 1 & \textit{X} (e.g., \texttt{CollectionClip}): cannot derive from sealed type \textit{X} \\
\texttt{CS7036} & 1 & There is no argument given that corresponds to the required formal parameter \textit{X} \\
\addlinespace \multicolumn{3}{@{}l}{\textit{Hygiene (H), 81 codes}}\\
\texttt{CS1003} & 28,571 & Syntax error, \textit{X} expected \\
\texttt{CS1002} & 15,943 & ; expected \\
\texttt{CS1525} & 4,524 & Invalid expression term \textit{X} \\
\texttt{CS1001} & 4,343 & Identifier expected \\
\texttt{CS1022} & 3,717 & Type or namespace definition, or end-of-file expected \\
\texttt{CS1513} & 2,852 & \} expected \\
\texttt{CS1044} & 2,374 & Cannot use more than one type in a for, using, fixed, or declaration statement \\
\texttt{CS0103} & 2,070 & The name \textit{X} does not exist in the current context \\
\texttt{CS1013} & 1,342 & Invalid number \\
\texttt{CS1529} & 1,138 & A using clause must precede all other elements defined in the namespace except extern alia \\
\texttt{CS1026} & 951 & ) expected \\
\texttt{CS0111} & 939 & Type \textit{X} already defines a member called \textit{X} with the same parameter types \\
\texttt{CS0101} & 820 & The namespace \textit{X} already contains a definition for \textit{X} \\
\texttt{CS1055} & 662 & An add or remove accessor expected \\
\texttt{CS1056} & 526 & Unexpected character \textit{X} \\
\texttt{CS0270} & 523 & Array size cannot be specified in a variable declaration (try initializing with a \textit{X} ex \\
\texttt{CS1519} & 497 & Invalid token \textit{X} in class, record, struct, or interface member declaration \\
\texttt{CS1012} & 472 & Too many characters in character literal \\
\texttt{CS1503} & 423 & Argument 1: cannot convert from \textit{X} to \textit{X} \\
\texttt{CS0116} & 403 & A namespace cannot directly contain members such as fields, methods or statements \\
\texttt{CS0650} & 369 & Bad array declarator: To declare a managed array the rank specifier precedes the variable' \\
\texttt{CS1514} & 339 & \{ expected \\
\texttt{CS8124} & 332 & Tuple must contain at least two elements. \\
\texttt{CS1010} & 328 & Newline in constant \\
\texttt{CS8803} & 169 & Top-level statements must precede namespace and type declarations. \\
\texttt{CS1014} & 167 & A get or set accessor expected \\
\texttt{CS1031} & 165 & Type expected \\
\texttt{CS1040} & 148 & Preprocessor directives must appear as the first non-whitespace character on a line \\
\texttt{CS0165} & 147 & Use of unassigned local variable \textit{X} \\
\texttt{CS0106} & 115 & The modifier \textit{X} is not valid for this item \\
\texttt{CS1526} & 111 & A new expression requires an argument list or (), [], or \{\} after type \\
\texttt{CS0595} & 107 & Invalid real literal. \\
\texttt{CS1041} & 71 & Identifier expected; \textit{X} is a keyword \\
\texttt{CS0263} & 55 & Partial declarations of \textit{X} must not specify different base classes \\
\texttt{CS8635} & 49 & Unexpected character sequence \textit{X} \\
\texttt{CS1585} & 47 & Member modifier \textit{X} must precede the member type and name \\
\texttt{CS1024} & 42 & Preprocessor directive expected \\
\texttt{CS0742} & 33 & A query body must end with a select clause or a group clause \\
\texttt{CS0443} & 32 & Syntax error; value expected \\
\texttt{CS8641} & 23 & \textit{X} cannot start a statement. \\
\texttt{CS0029} & 22 & Cannot implicitly convert type \textit{X} to \textit{X} \\
\texttt{CS0145} & 22 & A const field requires a value to be provided \\
\texttt{CS1018} & 17 & Keyword \textit{X} or \textit{X} expected \\
\texttt{CS0708} & 14 & \textit{X}: cannot declare instance members in a static class \\
\texttt{CS0178} & 13 & Invalid rank specifier: expected \textit{X} or \textit{X} \\
\texttt{CS0021} & 12 & Cannot apply indexing with [] to an expression of type \textit{X} \\
\texttt{CS0120} & 12 & An object reference is required for the non-static field, method, or property 'HerdAttract \\
\texttt{CS1073} & 12 & Unexpected token \textit{X} \\
\texttt{CS0687} & 10 & The namespace alias qualifier \textit{X} always resolves to a type or namespace so is illegal he \\
\texttt{CS7000} & 10 & Unexpected use of an aliased name \\
\texttt{CS0176} & 9 & Member \textit{X} cannot be accessed with an instance reference; quali \\
\texttt{CS0266} & 9 & Cannot implicitly convert type \textit{X} to \textit{X}. An explicit conversion exists \\
\texttt{CS1524} & 7 & Expected catch or finally \\
\texttt{CS1027} & 6 & \#endif directive expected \\
\texttt{CS0136} & 5 & A local or parameter named \textit{X} cannot be declared in this scope because that na \\
\texttt{CS1528} & 4 & Expected ; or = (cannot specify constructor arguments in declaration) \\
\texttt{CS0200} & 3 & Property or indexer \textit{X} cannot be assigned to -- it is read only \\
\texttt{CS0216} & 3 & The operator \textit{X} requires a matching operator \\
\texttt{CS0721} & 3 & \textit{X}: static types cannot be used as parameters \\
\texttt{CS1011} & 3 & Empty character literal \\
\texttt{CS1597} & 3 & Semicolon after method or accessor block is not valid \\
\texttt{CS1955} & 3 & Non-invocable member \textit{X} cannot be used like a me \\
\texttt{CS0034} & 2 & Operator \textit{X} is ambiguous on operands of type \textit{X} and \textit{X} \\
\texttt{CS0713} & 2 & Static class \textit{X} cannot derive from type \textit{X}. Static classes \\
\texttt{CS0722} & 2 & \textit{X}: static types cannot be used as return types \\
\texttt{CS1009} & 2 & Unrecognized escape sequence \\
\texttt{CS1501} & 2 & No overload for method \textit{X} takes 2 arguments \\
\texttt{CS1733} & 2 & Expected expression \\
\texttt{CS1950} & 2 & The best overloaded Add method \textit{X} for the collection initializer h \\
\texttt{CS0102} & 1 & The type \textit{X} already contains a definition for \textit{X} \\
\texttt{CS0230} & 1 & Type and identifier are both required in a foreach statement \\
\texttt{CS0272} & 1 & The property or indexer \textit{X} cannot be used in this context because the \\
\texttt{CS0664} & 1 & Literal of type double cannot be implicitly converted to type \textit{X}; use an \textit{X} suffix t \\
\texttt{CS0723} & 1 & Cannot declare a variable of static type \textit{X} \\
\texttt{CS1020} & 1 & Overloadable binary operator expected \\
\texttt{CS1515} & 1 & \textit{X} expected \\
\texttt{CS1553} & 1 & Declaration is not valid; use \textit{X} instead \\
\texttt{CS1646} & 1 & Keyword, identifier, or string expected after verbatim specifier: @ \\
\texttt{CS8180} & 1 & \{ or ; or =\textgreater{} expected \\
\texttt{CS8504} & 1 & Pattern missing \\
\texttt{CS8652} & 1 & The feature \textit{X} is currently in Preview and *unsupport \\
\end{longtable}
\endgroup
\twocolumn

\onecolumn
\section{Worked-Example Input: Stealth Pattern Description}
\label{app:stealth_md}
The complete goal-pattern description fed to the model for the worked example
(Section~\ref{sec:exemplar}). The model received this Markdown text verbatim and
nothing else --- no scene, assets, or asset names. The invented class names in the worked example (Section~\ref{sec:exemplar}), such as \texttt{GuardAI}, \texttt{AlarmSystem}, and \texttt{SafeZoneTrigger}, trace directly to terms in this text (Guards, Alarms, safe zone).

\begingroup\scriptsize
\begin{verbatim}
# Stealth

## Description

Stealth is the goal to move through a certain area and perform an action without being
detected.

## Overview

Sometimes favorable conditions in a game can be achieved by not having one's actions
noticed by other players. When this is the case, players have [Stealth](Stealth.md)
goals that force them to plan actions that minimize the risks of being noticed while
still completing the required actions.

[Stealth](Stealth.md) is a compound goal pattern using [Conceal](Conceal.md) together
with [Evade](Evade.md) with a secondary goal involving [Movement](Movement.md) or other
actions from the player, normally [Rescue](Rescue.md), [Traverse](Traverse.md),
[Delivery](Delivery.md), [Camping](Camping.md), or [Gain Ownership](GainOwnership.md)
(including gaining [Area Control](AreaControl.md) simply by being undetected in a
particular place). Designing the [Stealth](Stealth.md) goal consists not only of
choosing between the different design options of these patterns, but also determining
what player actions can reveal the players and what the [Tradeoffs](Tradeoffs.md) are
between the various [Risk/Reward](RiskReward.md) relations for each action in a given
context.

Longer [Stealth](Stealth.md) goals can be divided into parts that require short
[Stealth](Stealth.md) goals to be fulfilled in order to avoid [Guards](Guard.md) and
[Alarms](Alarms.md), short bursts of action to [Overcome](Overcome.md) enemy
[Units](Units.md) without them activating [Alarms](Alarms.md), and [Tension](Tension.md)
-filled moments when the best option for the player is to perform [No-Ops](No-Ops.md).
The complexity of [Stealth](Stealth.md) goals can be increased by letting
[Guards](Guard.md) have [Reconnaissance](Reconnaissance.md) goals so that players have
to take their [Movement](Movement.md) into consideration.

[Stealth](Stealth.md) is the goal of trying to [Conceal](Conceal.md) one's location
while having to move. [Stealth](Stealth.md) goals may require players to pace themselves
as quick [Movement](Movement.md) may have too high risks, and sometimes any action or
[Movement](Movement.md) may cause the goal to fail. [Stealth](Stealth.md) can thus
create [Tension](Tension.md) as players may have no [Freedom of
Choice](FreedomofChoice.md) except to perform [No-Op](No-Ops.md) actions to continue to
[Conceal](Conceal.md) themselves hoping not to be detected by opponents (which actually
represents a form of [Area Control](AreaControl.md)). The slow tempo and possible pauses
in completing [Stealth](Stealth.md) goalsgive players a chance to make use of [Strategic
Knowledge](StrategicKnowledge.md), for example the locations of [Alarms](Alarms.md),
making the game with the pattern have [Stimulated Planning](StimulatedPlanning.md).

Most cases of [Stealth](Stealth.md) rely on opponents having [Guard](Guard.md) or
[Reconnaissance](Reconnaissance.md) as [Preventing Goals](PreventingGoals.md), making
[Stealth](Stealth.md) and these goals[Excluding Goals](ExcludingGoals.md). Giving
players [Stealth](Stealth.md) goals combined with [Herd](Herd.md) goals increases the
chances of failure and may limit the[Right Level of
Difficulty](RightLevelofDifficulty.md) of the goals.

## Examples

* [Thief: The Dark Project](../game/thief-the-dark-project.md) and the other games in
the series exemplify a game using [Stealth](Stealth.md). The player is a master thief,
Garrett, who lives in a medieval fantasy world and performs his duties by relieving the
rich nobles of their riches. The main goal is to collect the valuable items, while the
secondary goal is to avoid being detected by the [Guards](Guard.md) while moving around
the [Level](Levels.md) s.
* Many children's' games are based on one person trying to find the other players while
at the same time trying to [Guard](Guard.md) an area that is a safe zone for the other
players. If the other players, by a combination of stealth and running, make it to the
safe zone they are home free and do not have to be the player guarding the safe zone in
the next game.

## Relations

### Instantiates

* [Evade](Evade.md)
* [Stimulated Planning](StimulatedPlanning.md)
* [Tension](Tension.md)
* [Movement](Movement.md)
* [Area Control](AreaControl.md)

### Modulates

* [Delivery](Delivery.md)
* [Rescue](Rescue.md)

### Instantiated by

* [Reconnaissance](Reconnaissance.md)

### Modulated by

* [Safe Havens](SafeHavens.md)
* [No-Ops](No-Ops.md)
* [Guard](Guard.md)
* [Alarms](Alarms.md)
* [Risk/Reward](RiskReward.md)
* [Tradeoffs](Tradeoffs.md)
* [Traverse](Traverse.md)
* [Gain Ownership](GainOwnership.md)
* [Camping](Camping.md)

### Potentially conflicting with

* [Herd](Herd.md)
\end{verbatim}
\endgroup

\section{Prompt Templates}
\label{app:prompts}

All prompts are reproduced verbatim from \texttt{src/prompts/} in the artifact repository.
Four placeholders are substituted at runtime. \texttt{<PATTERN\_ID>} becomes the goal-pattern
name, \texttt{<PATTERN\_MD>} the full Markdown description of that pattern
(see Appendix~\ref{app:stealth_md} for the Stealth example), \texttt{<IR\_JSON>}
the Step-1 IR generated by the IR-maker prompt, and \texttt{<METHOD>}
the IR conditioning level name
(\texttt{with\_schema\_free}, \texttt{with\_schema\_min}, or \texttt{with\_schema\_full}).

\subsection*{Coder prompts}

\noindent\textbf{Editor-style, \texttt{no\_schema}
(7B-Qwen2.5, 16B-DeepSeek, 22B-Codestral, 30B-Qwen3-Ed;
condition \texttt{no\_schema}):}
\begin{verbatim}
[pattern: <PATTERN_ID>]
[method: no_schema]

Generate a Unity Editor script that implements
the playable concept described below.
Output only raw C# code.

<PATTERN_MD>
\end{verbatim}

\noindent\textbf{Editor-style, \texttt{with\_schema}
(7B-Qwen2.5, 16B-DeepSeek, 22B-Codestral, 30B-Qwen3-Ed;
conditions \texttt{with\_schema\_free}/\texttt{min}/\texttt{full}):}
\begin{verbatim}
[pattern: <PATTERN_ID>]
[method: <METHOD>]

Generate a Unity Editor script that instantiates
a scene matching the following engine-specific
Intermediate Representation (IR). Thereafter,
you may refer to it as IR.
Output only raw C# code.

<IR_JSON>
\end{verbatim}

\noindent\textbf{Runtime-builder, \texttt{no\_schema}
(30B-Qwen3-Rt; condition \texttt{no\_schema}):}
\begin{verbatim}
[pattern: <PATTERN_ID>]
[method: no_schema_runtime]

Generate a single C# file containing a Unity MonoBehaviour called
RuntimeSceneBuilder that, in its Awake() method, programmatically
implements the playable concept described below.
Also define all gameplay MonoBehaviour classes in the same file.

Attach RuntimeSceneBuilder to an empty GameObject and press Play.
The scene must build itself at runtime — no Editor API, no
MenuItem, no UnityEditor namespace.
Output only raw C# code.

<PATTERN_MD>
\end{verbatim}

\noindent\textbf{Runtime-builder, \texttt{with\_schema}
(30B-Qwen3-Rt; conditions \texttt{with\_schema\_free}/\texttt{min}/\texttt{full}):}
\begin{verbatim}
[pattern: <PATTERN_ID>]
[method: <METHOD>]

Generate a single C# file containing a Unity MonoBehaviour called
RuntimeSceneBuilder that, in its Awake() method, programmatically
creates a scene matching the following engine-specific
Intermediate Representation (IR). Thereafter,
you may refer to it as IR.
Also define all gameplay MonoBehaviour classes in the same file.

Attach RuntimeSceneBuilder to an empty GameObject and press Play.
The scene must build itself at runtime — no Editor API, no
MenuItem, no UnityEditor namespace.
Output only raw C# code.

<IR_JSON>
\end{verbatim}

\subsection*{IR-maker prompts (Step~1, \texttt{with\_schema} conditions only)}

\noindent\textbf{\texttt{with\_schema\_free} (all five [model, generation mode]):}
\begin{verbatim}
[pattern: <PATTERN_ID>]
[method: with_schema_free]

Generate an engine-specific Intermediate Representation (IR) JSON for the
playable concept described below. Thereafter, you may refer to it as IR.
Output ONLY valid JSON. No extra text.

<PATTERN_MD>
\end{verbatim}

\noindent\textbf{\texttt{with\_schema\_min} (all five [model, generation mode]):}
\begin{verbatim}
[pattern: <PATTERN_ID>]
[method: with_schema_min]

Generate an engine-specific Intermediate Representation (IR) JSON for the
playable concept described below. Thereafter, you may refer to it as IR.
Output ONLY valid JSON. No extra text.

Required top-level fields:
  "scene"          — string
  "objects"        — [ { "id", "name", "type" }, ... ]
  "scripts"        — [ { "id", "object_id", "class_name" }, ... ]
  "params"         — {}
  "runtime_params" — { "<script_id>": { ... }, ... }
  "links"          — [ { "source", "target", "relation" }, ... ]
  "rules"          — [ { "id", "type", "description", "pattern", "evidence_type" }, ... ]

<PATTERN_MD>
\end{verbatim}

\noindent\textbf{\texttt{with\_schema\_full} (all five [model, generation mode]):}
\begin{verbatim}
[pattern: <PATTERN_ID>]
[method: with_schema_full]

Generate an engine-specific Intermediate Representation (IR) JSON for the
playable concept described below. Thereafter, you may refer to it as IR.
Output ONLY valid JSON. No extra text.

Follow the IR v0.2-runtime-evidence schema precisely.

Top-level fields (all required):
  "scene"          — string, scene identifier
  "objects"        — array of { "id", "name", "type" }
                     type in { "GameObject", "PrefabInstance", "PrefabAsset" }
  "scripts"        — array of { "id", "object_id", "class_name" }
                     one entry per component instance on one object
  "params"         — always {}
  "runtime_params" — object keyed by scripts[].id; values are flat { field: value } maps
  "links"          — array of { "source", "target", "relation", "evidence_type"? }
  "rules"          — array of { "id", "type", "description", "pattern",
                     "evidence_type", "confidence"? }

Hard constraints:
  1. Every scripts[].object_id MUST reference a real objects[].id (no dangling refs).
  2. Scripts are per-instance; no shared script entries across objects.
  3. Every entity must be listed explicitly in objects (no aggregate placeholders).
  4. Every rules[] entry MUST include evidence_type
     in { "direct_code", "scene_override", "inferred" }.

<PATTERN_MD>
\end{verbatim}

\end{document}